\documentclass[10pt,twocolumn,letterpaper]{article}

\usepackage[pagenumbers]{iccv} 

\newcommand{\red}[1]{{\color{red}#1}}

\usepackage{times}
\usepackage{epsfig}
\usepackage{graphicx}
\usepackage{amsmath}
\usepackage{amssymb}
\usepackage{booktabs}
\usepackage{multirow}
\usepackage{multicol}
\usepackage{longtable}
\usepackage[linesnumbered,ruled,vlined]{algorithm2e}

\newcommand{\ck}{\checkmark}
\newcommand{\bd}[1]{\textbf{#1}}

\newcommand{\ml}[2]{\multirow{#1}{*}{#2}}
\newcommand{\method}{tgGBC}
\newcommand{\Method}{TgGBC}

\definecolor{iccvblue}{rgb}{0.21,0.49,0.74}
\usepackage[pagebackref,breaklinks,colorlinks,allcolors=iccvblue]{hyperref}

\def\paperID{1071}
\def\confName{ICCV}
\def\confYear{2025}

\title{Accelerate 3D Object Detection Models via Zero-Shot Attention Key Pruning}

\author{
  \textbf{Lizhen Xu}$^{1}$,
  \textbf{Xiuxiu Bai}$^{1}$,
  \textbf{Xiaojun Jia}$^{2}$,
  \textbf{Jianwu Fang}$^{1}$,
  \textbf{Shanmin Pang}$^{1}$ \\  
  \textsuperscript{1}Xi'an Jiaotong University,
  \textsuperscript{2}Nanyang Technological University
}

\begin{document}
\maketitle

\begin{abstract}
Query-based methods with dense features have demonstrated remarkable success in 3D object detection tasks.
However, the computational demands of these models, particularly with large image sizes and multiple transformer layers, pose significant challenges for efficient running on edge devices.
Existing pruning and distillation methods either need retraining or are designed for ViT models, which are hard to migrate to 3D detectors.
To address this issue, we propose a zero-shot runtime pruning method for transformer decoders in 3D object detection models.
The method, termed \textbf{\method{}} (\textbf{t}rim keys \textbf{g}radually \textbf{G}uided \textbf{B}y \textbf{C}lassification scores), systematically trims keys in transformer modules based on their importance.
We expand the classification score to multiply it with the attention map to get the importance score of each key and then prune certain keys after each transformer layer according to their importance scores.
Our method achieves a 1.99x speedup in the transformer decoder of the latest ToC3D model, with only a minimal performance loss of less than 1\%.
Interestingly, for certain models, our method even enhances their performance.
Moreover, we deploy 3D detectors with \method{} on an edge device, further validating the effectiveness of our method.
The code can be found at \url{https://github.com/iseri27/tg\_gbc}.
\end{abstract}

\section{Introduction}
\label{sec:intro}

\begin{figure}
    \centering
    \includegraphics[width=\linewidth]{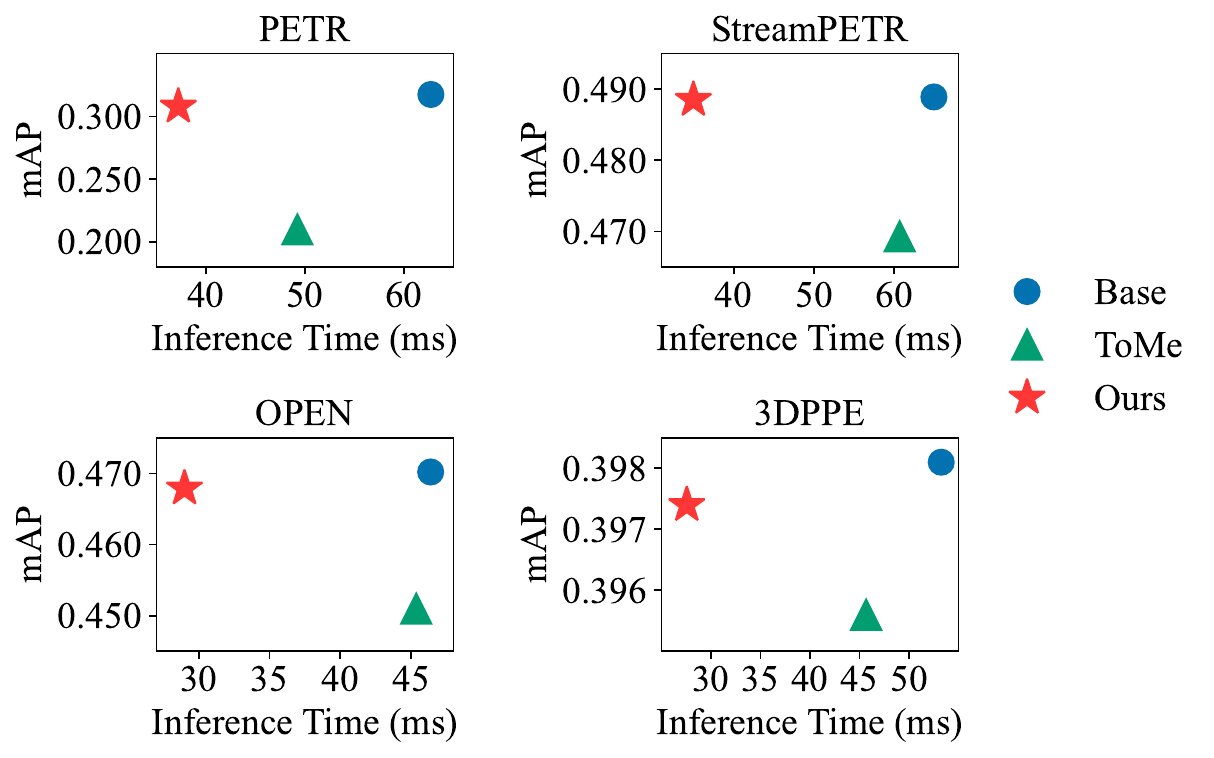}
    \vskip -0.1in
    \caption{
    The mAP and inference time of models, comparing with applying \method{} and ToMe~\cite{tome}.
    }
    \label{fig:compare_base_gbc_tome}
    \vskip -0.15in
\end{figure}

Vision-based 3D object detection is a crucial technique that empowers intelligent devices to perceive and interpret the real world precisely.
In recent years, query-based models \cite{detr3d,petr,petrv2,focalPetr,streampetr,raydn,open,toc3d,mbev,mv2d,bevdepth,bevformer,bevformerv2} have risen to prominence as the state-of-the-art paradigm in the field of 3D object detection.
These methodologies typically employ a CNN or ViT backbone \cite{resnet,vovnetv2,vit,eva02} for image feature extraction, followed by sophisticated processing through a multi-layer transformer decoder.
Based on their distinct approaches to feature map processing and attention mechanism implementation, these methods can be fundamentally classified into two categories: \textbf{dense methods} and \textbf{sparse methods}.
Dense methods facilitate comprehensive global interactions across entire feature maps \cite{petr,petrv2,focalPetr,streampetr,3dppe,open}, enabling holistic information integration.
In contrast, sparse methods accelerate model convergence by focusing on a carefully selected subset of feature maps~\cite{detr3d,sparsebev,far3d,bevformer,bevformerv2}, often leveraging deformable attention mechanisms~\cite{deformabledetr} to improve computational efficiency.
Despite significant progress in sparse methodologies, dense methods continue to play a pivotal role in 3D object detection, largely attributed to their superior detection accuracy (see detailed discussion in \cref{sec:appendix:3d_detectors:advantages_over_sparse_methods}).
The sustained importance of dense approaches is further strengthened through ongoing advancements, where researchers are actively proposing and implementing novel methodologies and architectural refinements \cite{open,toc3d}.

Nevertheless, dense methods also present a notable limitation: their characteristic global feature interactions incur substantial computational overhead, requiring dedicated optimization strategies like pruning~\cite{lth,sixteen,svit,svite,wdpruning,layerdrop} or distillation~\cite{distilling_learning,task_aware_distill,detrdistill} for effective deployment on resource-constrained edge platforms.
In this context, retraining-free pruning approaches have recently emerged as a promising research direction, attracting considerable attention from the community~\cite{xpruner,fast_post_traing_pruning,optin,zero_tprune,ats,tome}.
Based on the phase in which pruning methods are applied, they can be broadly categorized into two distinct types: \textbf{static pruning} and \textbf{runtime pruning}. 
Static pruning methods \cite{fast_post_traing_pruning,optin} typically operate by executing the model on a small-scale subset of data to identify and preserve critical parameters. While these approaches effectively reduce both model parameters and GPU memory consumption during runtime, they inherently require model execution to conduct the parameter search.
The efficiency and efficacy of static pruning are influenced by several factors, including the model's architectural complexity, its scale, and the representativeness of the subset data used for pruning.
Moreover, each static pruning method is specifically designed for a particular model or class of methods.
Therefore, static pruning methods designed for image classification tasks\cite{optin} cannot be directly transferred to 3D detection tasks, limiting their universality.

Runtime pruning methods integrate pruning layers between specific model modules without altering the original model parameters, thereby enabling dynamic pruning during inference \cite{ats,tome,zero_tprune}.
This approach provides a plug-and-play solution that incurs no additional training costs while significantly accelerating model inference, making it a zero-overhead optimization strategy.
Unlike static pruning, runtime pruning eliminates the need for retraining, parameter search, or reliance on a sub-dataset.
Users only need to integrate the pruning layer into the model and execute the inference procedure as usual.

To accelerate model inference speed without retraining, we propose \method{}, a runtime pruning method that progressively removes keys from transformer modules based on their importance.
3D object detectors usually adopt stacked transformer layers, each comprising a self-attention and a cross-attention module.
During inference, these detectors determine which predictions to retain based on the classification scores, as higher scores indicate more accurate bounding boxes and attributes. We integrate our \method{} module between two transformer layers to reduce the number of keys.
This module takes as input the classification scores from a CNN head and the attention map produced by the cross-attention module.
By multiplying the classification scores with the attention map and then computing their sum, we transfer the quality of the classification scores to each key. 
The least important $r$ keys are then pruned, significantly optimizing computation.
As shown in \cref{fig:compare_base_gbc_tome}, integrating \method{} into popular 3D detection models results in nearly a $2\times$ speedup in transformer decoder inference while incurring only a minimal drop in mAP.

To summarize, our contributions are as follows:
\begin{itemize}
    \item We propose \method{}, a zero-shot pruning technique that accelerates dense transformer decoders in 3D object detectors. To the best of our knowledge, this is the first work to explore zero-shot, retraining-free pruning for 3D object detection models.
    \item We calculate the importance of each key using the classification scores and attention maps that are inherently generated within the transformer decoder without introducing any additional parameters.
    \item Extensive experimental results show that \method{} accelerates the transformer decoder of  dense 3D detectors by nearly $2\times$ with minimal performance degradation. Furthermore, when deploying FocalPETR~\cite{focalPetr} and StreamPETR~\cite{streampetr} on an edge device with \method{}, we respectively achieve  $1.18\times $ and $1.19\times$ speedup in inference, demonstrating its efficiency in real-world scenarios.
\end{itemize}

\section{Related Works}
\label{sec:related_works}

\subsection{Query-based 3D Object Detection Methods}

Transformer~\cite{transformer}, as a prominent attention mechanism, has been extensively adopted in object detection tasks.
DETR~\cite{detr} is the first model to successfully integrate transformer architecture into 2D object detection, utilizing learnable queries for object prediction.
Building upon this framework, the query-based paradigm has demonstrated remarkable success in advancing 3D object detection methodologies.
PETR \cite{petr} generates position-aware 3D features to help detect objects.
FocalPETR \cite{focalPetr} uses a 2D auxiliary task to select more confident priors.
3DPPE \cite{3dppe} introduces 3D point positional encoding by using a depth predictor.
OPEN \cite{open} proposes an object-wise position embedding to inject object-wise depth information into the network.
M-BEV \cite{mbev} introduces a generic masked BEV perception framework designed to handle situations where one or more cameras fail, which enhances the safety and robustness of perception algorithms.
ToC3D \cite{toc3d} proposes a token compression backbone to speed up the model's inference.

Despite their varying architectural designs, the aforementioned 3D object detection methods maintain a consistent framework characterized by an image backbone coupled with a multi-layer transformer decoder.

\begin{figure*}[t]
    \centering
    \includegraphics[width=0.95\textwidth]{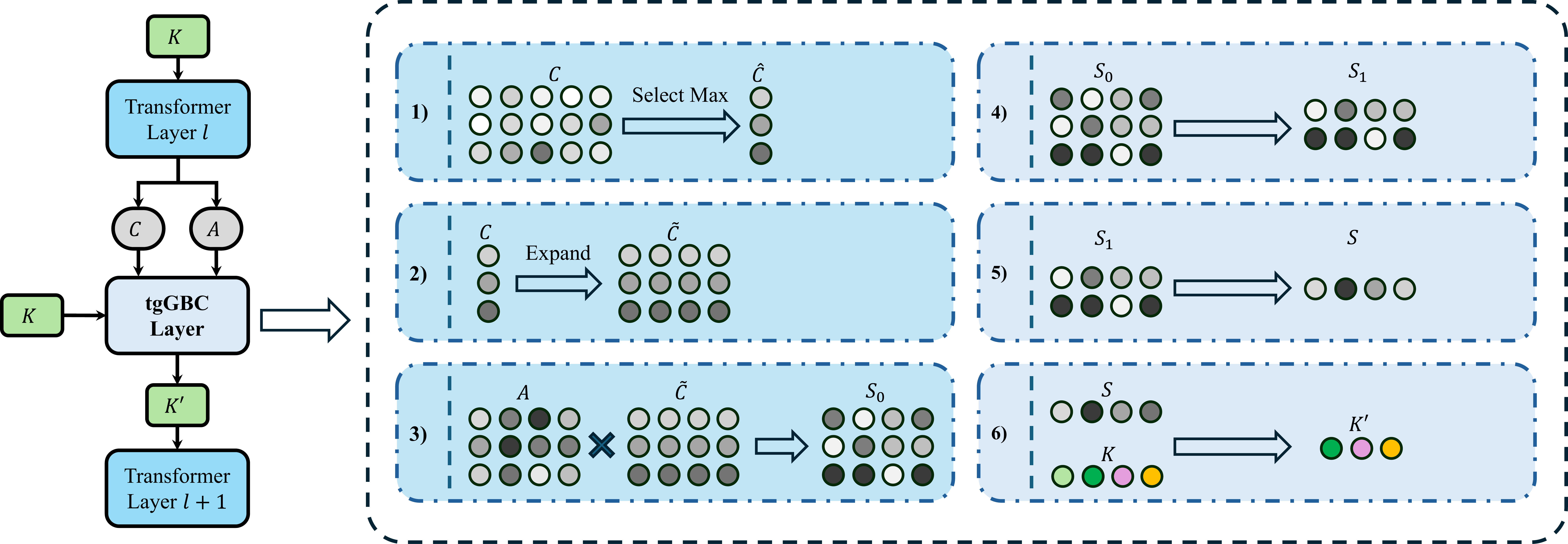}
    \caption{
    \Method{} layer is inserted after the first $n$ transformer decoder layers, taking classification scores $C$, attention weights $A$ and keys $K$ as input.
    Inside the \method{} layer,
    1) we select the maximum of each classification vector, resulting in $\hat C\in\mathbb R^{N_q}$;
    2) repeat and expand $\hat C$ to a shape of $\mathbb R^{N_q\times N_k}$, got $\tilde C$;
    3) calculate $S_0=A\odot \tilde C\in\mathbb R^{N_q\times N_k}$;
    4) select $k$ lines of $S_0$ with highest $k$ classification scores, got $S_1$;
    5) conduct sum of $S_1$ along column, got key importance $S$;
    6) prune $r$ keys with the lowest importance value.
    }
    \label{fig:tgGBC}
    \vskip -0.1in
\end{figure*}

\subsection{Retraining-free Pruning for Transformers}
\label{subsec:related_works:pruning_methods}

Retraining-free pruning methodologies have recently gained significant attention in the research community.
Among these, static pruning techniques \cite{optin,fast_post_traing_pruning} have shown particular promise by enabling model optimization before deployment, typically utilizing small-scale datasets for parameter adjustment.

In parallel development, runtime pruning approaches insert a pruning layer inside the model.
ATS \cite{ats} uses classification tokens---a special structure in ViT models---as pruning criterion.
ToMe \cite{tome} divides tokens into two groups and uses a bipartite matching algorithm to pair them one-to-one, merging the top-$r$ pairs with the highest $r$ similarity.
Notably, Zero-TPrune \cite{zero_tprune} employs an extended Markov algorithm, treating the attention map as the state transition matrix of a Markov chain and using the converged state as the pruning criterion for tokens.

However, these methods face significant challenges when adapted to 3D detection tasks.
In query-based 3D detection models, the use of panoramic cameras and high-resolution image inputs leads to a substantial increase in token number compared to ViT-based models.
For instance, while ViT-based models \cite{vit, deit,eva02} typically process 1,024 tokens at a resolution of $448\times 448$, 3D object detection models must handle a significantly larger token space---4,224 tokens even at a minimal resolution of $704\times 256$.
This dramatic increase in token count severely impacts the computational efficiency of methods like ToMe, which relies on similarity matrix calculations with quadratic time complexity.
Additionally, Zero-TPrune's architectural design, though effective for ViT models, is fundamentally limited by its dependence on square attention matrices.
In ViT architectures, tokens simultaneously serve as \textit{query}, \textit{key}, and \textit{value} within attention modules.
In contrast, 3D object detection models utilize distinct \textit{query} and \textit{key}, resulting in the generation of non-square attention maps, making it impossible to adapt Zero-TPrune to 3D detectors.

To address these challenges, we introduce a runtime pruning module optimized for 3D object detection tasks. 
\section{Method}
\label{sec:method}

\subsection{Preliminary and Feasibility}
\label{subsec:method:pre-and-feasibility}

Before delving into our approach, it is essential to understand why pruning keys is feasible.

Consider a multi-head attention mechanism with $N_h$ heads and an embedding dimension of $E$.
A key requirement is that \( E \) must be an integer multiple of \( N_h \) to ensure that the embedding can be evenly partitioned across all attention heads.
We define the dimensionality of each attention head as \( E_h = E / N_h \).
The multi-head attention mechanism begins by projecting the input queries, keys, and values using learnable parameter matrices $W^Q, W^K, W^V\in\mathbb R^{E\times E}$, formulated as:
\begin{equation}
    \begin{aligned}
        Q \leftarrow QW^Q,~~K \leftarrow KW^K,~~V \leftarrow VW^V
    \end{aligned}
    \label{eq:mha:op:qkv_proj}
\end{equation}
where $Q\in\mathbb R^{N_q\times E}$, $K, V\in\mathbb R^{N_k\times E}$ represent \textit{query}, \textit{key} and \textit{value} matrices respectively.

To enable multi-head attention, the projected matrices $Q$ is split along the embedding dimension to obtain head-specific representations \( q^h \):
\begin{equation}
    \begin{aligned}
        q^h & = Q_{:,h\times E_h:(h+1)\times E_h} \in\mathbb R^{N_q\times E_h}
    \end{aligned}
    \label{eq:mha:op:qkv_split}
\end{equation}
$k^h$ and $v^h$ are obtained similarly. For each attention head $h$, the attention map matrix $A^h$ is computed using the scaled dot product:
\begin{equation}
    \begin{aligned}
        A^h & = \text{Softmax}\left( \cfrac{q^h\times (k^h)^\mathrm T}{\sqrt{E}} \right)\in\mathbb R^{N_q\times N_k}
    \end{aligned}
    \label{eq:mha:op:attn_weights}
\end{equation}
The attention map $A^h$ is then multiplied by $v^h$ to obtain the head-specific output $O^h$: $O^h = A^h \times v^h\in\mathbb R^{N_q\times E_h}$.
The final output $O \in \mathbb{R}^{N_q \times E}$ of the attention module is obtained by concatenating the outputs from all attention heads:
\begin{equation}
O  = \mathop{\text{Concat}}\limits_{h\in\{1,\cdots, N_h\}}\left( O^h \right)\times W^O, O\in\mathbb R^{N_q\times E}
    \label{eq:mha:op:attn_out}
\end{equation}
where $W^O\in\mathbb R^{E\times E}$ is a learnable parameter.

As shown in Figure \ref{fig:3d_detector} (in \cref{sec:appendix:3d_detectors:overall_arch}), 3D detectors typically include both self-attention and cross-attention modules within each transformer layer.
The self-attention module takes pre-defined queries as $Q$, $K$ and $V$, and its output is then used as $Q$ in the subsequent cross-attention module.
Notably, in 3D object detection models, $V$ is always identical to $K$.

From \cref{eq:mha:op:qkv_proj}-\cref{eq:mha:op:attn_out}, we observe that the shape of the attention output is independent of $N_k$.
Therefore,  for an attention module with learnable parameters $W^Q, W^K, W^V, W^O\in\mathbb R^{E\times E}$, the original parameters remain valid even when modifying $N_q$ and $N_k$, as long as the following two conditions are met:
1) the embedding dimension $E$ remains unchanged to match the projection parameters $W^Q, W^K, W^V$ and $W^O$;
2) the number of keys and values remains equal.

Thus, it is feasible to dynamically prune keys in dense models with global attention during runtime without modifying the model parameters. However, it should be noted that key pruning may not be applicable to models utilizing sparse attention mechanisms, such as deformable attention or flash attention, since these approaches inherently lack an attention map \cite{sparsebev,raydn,detr3d,far3d,deformabledetr}.

\subsection{Pruning Criterion}
\label{subsec:pruning_criterion}

As highlighted in GPQ \cite{gpq}, predictions with the highest classification scores are retained as final outputs.
Consequently, keys that contribute minimally to these high-confidence predictions can be considered unimportant and pruned without significantly affecting the final results.

According to \cref{eq:mha:op:qkv_proj}-\cref{eq:mha:op:attn_out}, it is clear that the number of final predictions is determined by the number of queries $N_q$ rather than the number of keys $N_k$.
This implies that pruning should first target unimportant queries—those whose predictions are less likely to be selected in the final results. Fortunately, query importance can be directly assessed using classification scores \cite{gpq}.

Once unimportant queries are identified, the next step is to find which keys are most relevant to them.
The attention map entry $A_{i,j}$ directly represents the correlation between query $q_i$ and key $k_j$.
Thus, a natural pruning criterion is to compute the importance of each key as the sum of the product of classification scores and attention weights. Keys with the lowest importance values can then be pruned, reducing computational cost while preserving model performance.

\subsection{Pruning Process}

The pruning process is illustrated in \cref{fig:tgGBC}.
Before pruning, we first need to obtain classification scores $C\in\mathbb R^{N_q\times N_C}$ and attention weights $\{A^h\}$ from the previous transformer decoder layer, where $N_C$ represents the number of object classes to be predicted.
\Method{} begins by selecting the maximum value from each classification vector $C_{i}$, yielding $\hat C\in\mathbb R^{N_q}$:
\begin{equation}
\hat C_i=  \max\limits_j C_{i, j},  ~\forall i\in\{1, 2, \cdots, N_q\}
    \label{eq:selec_cls_maximum}
\end{equation}
Then, $\hat C$ is repeated $N_k$ times, resulting in the expanded classification scores $\tilde{C}\in\mathbb R^{N_q\times N_k}$:
\begin{equation}
    \begin{aligned}
        \tilde{C}_{i,m} & = \hat C_i, ~\forall m\in\{1, \cdots, N_k\}
    \end{aligned}
    \label{eq:expand_cls_scores}
\end{equation}

Next, we conduct element-wise multiplication between the averaged attention map $A$ and the expanded classification scores $\tilde{C}$:
\begin{equation}
    \begin{aligned}
        S_0 = A\odot \tilde C,~~A = \cfrac{1}{N_h}\sum\limits_{i=1}A_i
    \end{aligned}
    \label{eq:a_dot_c}
\end{equation}
where $S_0\in\mathbb R^{N_q\times N_k}$, its $i$-th row, $(S_0)_i$, corresponds to the classification score $\hat C_i$.
We then select the top-$k$ rows with the highest classification scores, obtaining $S_1\in\mathbb R^{k\times N_k}$:
\begin{equation}
    \begin{aligned}
        S_1 = (S_0)_{\Phi},~~\Phi = \{i | \max\limits_j C_{i,j} \geq c_k \}
    \end{aligned}
    \label{eq:select_k_lines}
\end{equation}
where $c_k$ represents the $k$-th largest value in $\hat C$.

Finally, we  compute the sum along the columns of $S_1$, yielding the token importance scores $S\in\mathbb R^{N_k}$:
\begin{equation}
    \begin{aligned}
        S_j = \sum\limits_{i=1}^k (S_1)_{i,j}
    \end{aligned}
    \label{eq:sum}
\end{equation}
Once the importance scores $S$ are obtained, we prune the keys with the lowest importance values.

Assuming the transformer decoder has $L$ layers and takes $N_k$ keys as input, our goal is to prune a total of $r$ keys.
To achieve this, we remove $\left\lfloor\frac{r}{n}\right\rfloor$ keys per layer, where $1\leq n\leq N_k-1$.
Unlike ToMe, which relies on bipartite matching, our approach allows pruning all $r$ keys within a single layer, even when $r$ is close to 90\% of $N_k$.

The analysis of the cost of \method{} can be found in \cref{sec:appendix:analysis}.

\section{Experiments}
\label{sec:experiments}

\newcommand{\petr}{\begin{tabular}[c]{@{}c@{}}PETR~\cite{petr}\\(ECCV 2022)\end{tabular}}
\newcommand{\focalpetr}{\begin{tabular}[c]{@{}c@{}}FocalPETR~\cite{focalPetr}\\(IEEE TIV 2023)\end{tabular}}
\newcommand{\streampetr}{\begin{tabular}[c]{@{}c@{}}StreamPETR~\cite{streampetr}\\(ICCV 2023)\end{tabular}}
\newcommand{\threeDppe}{\begin{tabular}[c]{@{}c@{}}3DPPE~\cite{3dppe}\\(ICCV 2023)\end{tabular}}
\newcommand{\mvTwoD}{\begin{tabular}[c]{@{}c@{}}MV2D~\cite{mv2d}\\(ICCV 2023)\end{tabular}}
\newcommand{\mbev}{\begin{tabular}[c]{@{}c@{}}M-BEV~\cite{mbev}\\(AAAI 2024)\end{tabular}}
\newcommand{\open}{\begin{tabular}[c]{@{}c@{}}OPEN~\cite{open}\\(ECCV 2024)\end{tabular}}
\newcommand{\tocThreeD}{\begin{tabular}[c]{@{}c@{}}ToC3D~\cite{toc3d}\\(ECCV 2024)\end{tabular}}

\begin{table*}[ht]
\centering
\resizebox{\textwidth}{!}{
\begin{tabular}{
    cc@{\hspace{5pt}}c|
    c@{\hspace{5pt}}c|
    c@{\hspace{5pt}}c|
    c@{\hspace{5pt}}c@{\hspace{5pt}}c@{\hspace{5pt}}c@{\hspace{5pt}}c|
    c@{\hspace{5pt}}c
}
\toprule
\bd{Model}
& \bd{Backbone}
& \bd{ImageSize}
& $N_k$
& \bd{$r$}
& \bd{mAP} $\uparrow$
& \bd{NDS} $\uparrow$
& \bd{mATE} $\downarrow$
& \bd{mASE} $\downarrow$
& \bd{mAOE} $\downarrow$
& \bd{mAVE} $\downarrow$
& \bd{mAAE} $\downarrow$
& \bd{Inf. Time (ms)} $\downarrow$
& \bd{Dec. Time (ms)} $\downarrow$ \\ \midrule
\ml{4}{\petr}      & \ml{2}{ResNet50} & \ml{2}{1408x512}& \ml{2}{16896}& -     & 31.74\% & 0.3669 & 0.8392 & 0.2797 & 0.6145 & 0.9521 & 0.2322 & 131.94           & 47.09           \\
                   &                  &                 &              & 12000 & 30.78\% & 0.3579 & 0.8540 & 0.2832 & 0.6168 & 0.9703 & 0.2354 & 112.88(-14.45\%) & 28.58(-39.31\%) \\ \cline{2-14} 
                   & \ml{2}{VovNet}   & \ml{2}{1600x640}& \ml{2}{24000}& -     & 40.45\% & 0.4517 & 0.7287 & 0.2706 & 0.4485 & 0.8399 & 0.2178 & 281.05           & 62.72           \\
                   &                  &                 &              & 18000 & 39.53\% & 0.4432 & 0.7482 & 0.2720 & 0.4538 & 0.8539 & 0.2167 & 254.18(-9.56\%)  & 37.22(-40.66\%) \\ \midrule
\ml{4}{\focalpetr} & \ml{2}{ResNet18} & \ml{2}{1408x512}& \ml{2}{16896}& -     & 32.24\% & 0.3566 & 0.7684 & 0.2803 & 0.6530 & 1.0942 & 0.3447 & 80.87            & 44.26           \\
                   &                  &                 &              & 12000 & 31.70\% & 0.3530 & 0.7767 & 0.2806 & 0.6503 & 1.0935 & 0.3466 & 62.10(-23.21\%)  & 26.01(-41.23\%) \\ \cline{2-14} 
                   & \ml{2}{VovNet}   & \ml{2}{800x320} & \ml{2}{6000} & -     & 42.36\% & 0.4716 & 0.6572 & 0.2658 & 0.4582 & 0.8050 & 0.2159 & 99.14            & 24.65           \\
                   &                  &                 &              & 3000  & 42.38\% & 0.4719 & 0.6560 & 0.2663 & 0.4572 & 0.8045 & 0.2157 & 89.66(-9.56\%)   & 17.08(-30.71\%) \\ \midrule
\ml{4}{\streampetr}& \ml{2}{ResNet50} & \ml{2}{704x256} & \ml{2}{4224} & -     & 38.01\% & 0.4822 & 0.6781 & 0.2763 & 0.6401 & 0.2831 & 0.2007 & 60.52            & 31.10           \\
                   &                  &                 &              & 2000  & 37.93\% & 0.4817 & 0.6787 & 0.2758 & 0.6390 & 0.2844 & 0.2016 & 51.68(-14.61\%)  & 24.15(-22.35\%) \\ \cline{2-14} 
                   & \ml{2}{VovNet}   & \ml{2}{1600x640}& \ml{2}{24000}& -     & 48.89\% & 0.5732 & 0.6096 & 0.2601 & 0.3882 & 0.2603 & 0.1944 & 288.07           & 64.93           \\
                   &                  &                 &              & 21000 & 48.55\% & 0.5730 & 0.6033 & 0.2626 & 0.3771 & 0.2611 & 0.1941 & 254.73(-11.57\%) & 34.98(-46.13\%) \\ \midrule
\ml{2}{\threeDppe} & \ml{2}{VovNet}   & \ml{2}{800x320} & \ml{2}{6000} & -     & 39.81\% & 0.4460 & 0.7040 & 0.2699 & 0.4951 & 0.8438 & 0.2177 & 125.13           & 53.25           \\
                   &                  &                 &              & 3000  & 39.74\% & 0.4449 & 0.7057 & 0.2707 & 0.4956 & 0.8465 & 0.2202 & 96.65(-22.76\%)  & 27.56(-48.24\%) \\ \midrule
\ml{2}{\mvTwoD}    & \ml{2}{ResNet50} & \ml{2}{1408x512}& \ml{2}{-}    & -     & 44.92\% & 0.5399 & 0.6246 & 0.2657 & 0.3840 & 0.4009 & 0.1722 & 416.61           & 96.58           \\
                   &                  &                 &              & 50\%  & 44.11\% & 0.5384 & 0.6248 & 0.2657 & 0.3844 & 0.4024 & 0.1717 & 388.58(-6.73\%)  & 70.83(-26.66\%) \\ \midrule
\ml{2}{\mbev}      & \ml{2}{VovNet}   & \ml{2}{800x320} & \ml{2}{12000}& -     & 35.14\% & 0.4640 & 0.7300 & 0.2717 & 0.4980 & 0.4324 & 0.1845 & 178.87           & 55.14           \\
                   &                  &                 &              & 6000  & 34.37\% & 0.4557 & 0.7520 & 0.2744 & 0.5099 & 0.4376 & 0.1872 & 151.00(-15.58\%) & 33.96(-38.41\%) \\ \midrule
\ml{6}{\open}      & \ml{2}{ResNet50} & \ml{2}{704x256} & \ml{2}{4224} & -     & 47.02\% & 0.5657 & 0.5676 & 0.2702 & 0.4221 & 0.2321 & 0.2019 & 77.25            & 33.63           \\
                   &                  &                 &              & 2000  & 46.85\% & 0.5637 & 0.5682 & 0.2705 & 0.4311 & 0.2325 & 0.2031 & 65.18(-15.62\%)  & 24.23(-27.95\%) \\ \cline{2-14} 
                   & \ml{2}{VovNet}   & \ml{2}{800x320} & \ml{2}{6000} & -     & 52.07\% & 0.6128 & 0.5250 & 0.2566 & 0.2811 & 0.2148 & 0.1982 & 118.55           & 29.35           \\
                   &                  &                 &              & 3000  & 52.09\% & 0.6129 & 0.5249 & 0.2569 & 0.2808 & 0.2146 & 0.1986 & 111.40(-6.03\%)  & 23.26(-20.75\%) \\ \cline{2-14} 
                   & \ml{2}{ResNet101}& \ml{2}{1408x512}& \ml{2}{16896}& -     & 51.80\% & 0.6043 & 0.5314 & 0.2679 & 0.3457 & 0.2095 & 0.1922 & 196.55           & 46.39           \\
                   &                  &                 &              & 12000 & 51.57\% & 0.6019 & 0.5368 & 0.2726 & 0.3493 & 0.2102 & 0.1905 & 178.35(-9.31\%)  & 28.99(-37.51\%) \\ \midrule
\ml{2}{\tocThreeD} & \ml{2}{ToC3DViT} & \ml{2}{1600x800}& \ml{2}{30000}& -     & 54.20\% & 0.6187 & 0.5589 & 0.2571 & 0.2716 & 0.2353 & 0.2007 & 863.47           & 77.92           \\
                   &                  &                 &              & 27000 & 53.43\% & 0.6113 & 0.5842 & 0.2589 & 0.2744 & 0.2370 & 0.2038 & 817.69(-5.30\%)  & 39.17(-49.73\%) \\
\bottomrule
\end{tabular}
}
\caption{Pruning results for different models.
Results shown in the table all use a setting of $n$ = 2 and $k$ = 175 when \method{} is applied.
The column ``$N_k$'' indicates the number of keys.
We report inference time of the whole model as shown in column ``Inf. Time'' and the time of the transformer decoder only as shown in the column ``Dec. Time''.
The speed is tested on a single RTX3090.
Due to MV2D's special design, its number of keys is floating, hence, we prune 50\% of keys according to the current number of keys.
}
\label{tab:results-performance}
\vskip -0.15in
\end{table*}

\begin{figure*}[t]
    \centering
    \includegraphics[width=0.99\linewidth]{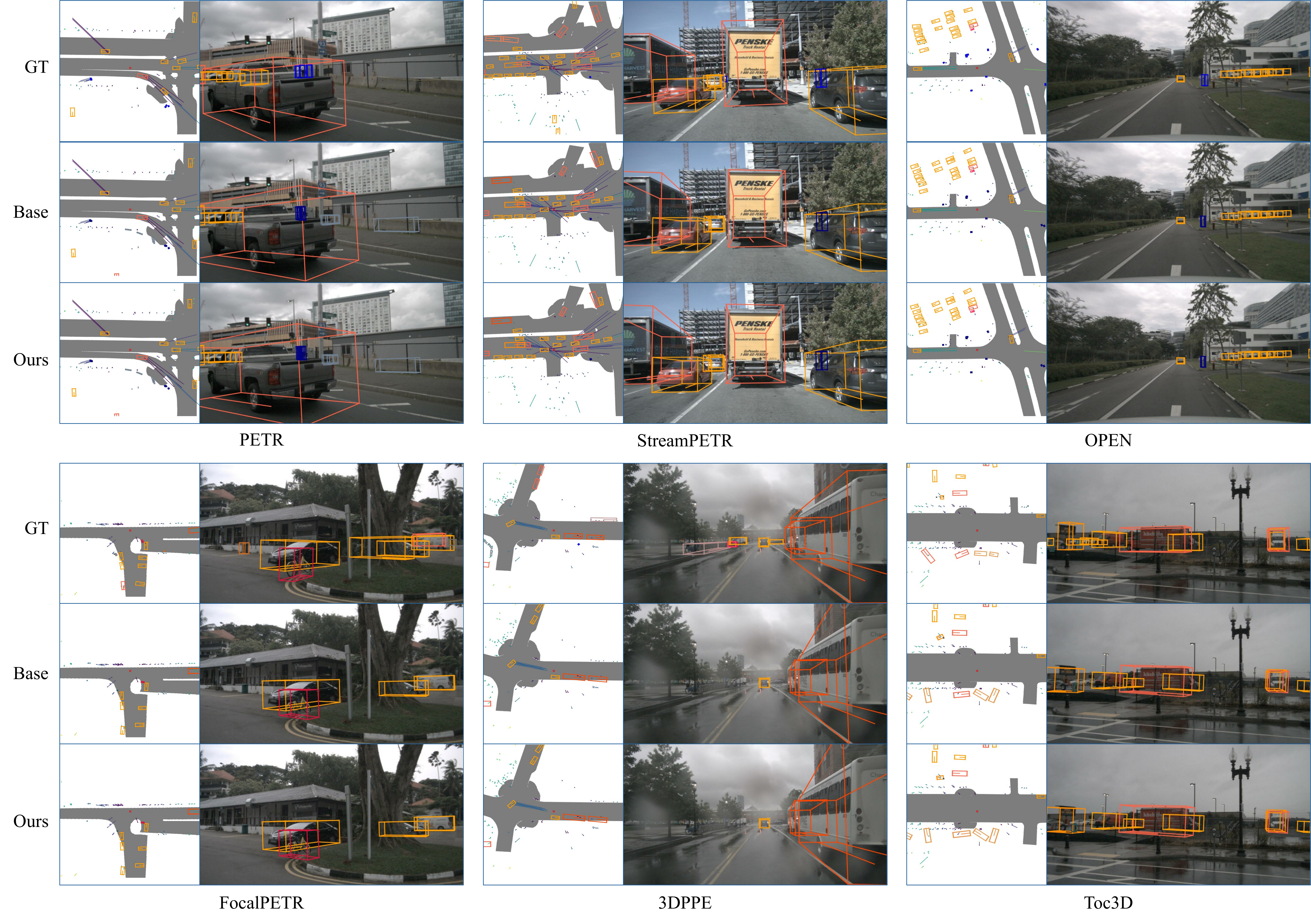}
    \vskip -0.1in
    \caption{Visualization results for different models with top-down view(left for each sub-figure) and camera view.}\label{fig:results_visualization}
    \vskip -0.1in
\end{figure*}

\subsection{Experimental Setup}


Our experiments are conducted on the nuScenes \cite{nuscenes} dataset, and we use evaluation metrics established by this dataset.
The primary evaluation metric is mean Average Precision (mAP), which serves as the fundamental measure of detection accuracy. In addition to spatial localization, 3D object detection requires precise prediction of object characteristics, including orientation, velocity, and specific attributes. To comprehensively assess these capabilities, nuScenes introduces five complementary metrics collectively referred to as mTP: mean Average Translation Error (mATE), mean Average Scale Error (mASE), mean Average Orientation Error (mAOE), mean Average Velocity Error (mAVE), and mean Average Attribute Error (mAAE).

To offer a holistic evaluation framework, nuScenes proposes the nuScenes Detection Score (NDS), 
providing a comprehensive assessment of model performance across all critical detection aspects.
The NDS is computed using the formula:
    $\text{NDS} = \cfrac{1}{5}\left[
        5\text{mAP} + \sum\limits_{\text{mTP}}\left(
            1-\min(1, \text{mTP})
        \right)
    \right],$
where $\text{mTP}\in\{\text{mATE}, \text{mASE}, \text{mAOE}, \text{mAVE}, \text{mAAE}\}$.

To validate the effectiveness and efficiency of \method{}, we perform experiments on eight 3D advanced detectors:
PETR \cite{petr}, FocalPETR \cite{focalPetr}, StreamPETR \cite{streampetr}, OPEN \cite{open}, ToC3D \cite{toc3d}, 3DPPE \cite{3dppe}, MV2D \cite{mv2d} and M-BEV \cite{mbev}.
All experiments are conducted under their original configurations to ensure fair and consistent comparisons.

\subsection{Effectiveness Assessment}
\label{sec:experiments:results_performance}

The primary challenge for retraining-free pruning methods is maintaining model performance while improving efficiency.
\cref{tab:results-performance} reports the performance and inference time of various models after pruning about 50\%-90\% of keys.
The maximum number of keys that can be pruned varies across different models.
To ensure a fair and clear comparison, we use a standardized experimental setup of \(n = 2\) and \(k = 175\).
While these results may not represent the optimal configurations for every model, they provide a clear evaluation of the effectiveness of \method{}.

For 3DPPE with VovNet as the backbone and an image size of  $800\times 320$, pruning 3000 keys (50\% of the original 6000 keys) results in an mAP of 39.74\%, with only a 0.07\% drop.
Meanwhile, inference time saves by 22.76\% ($1.29\times$ faster) for the whole model and 48.24\% ($1.93\times$ faster) for the transformer decoder.
Interestingly, for certain models such as FocalPETR-vov-800x320 and OPEN-vov-800x320, applying \method{} leads to a slight increase in mAP, suggesting potential benefits beyond just efficiency improvements.

For models with large image sizes, such as StreamPETR-vov-1600x640, \method{} successfully prunes 21,000 out of 24,000 keys (87.5\%), resulting in an 11.57\% reduction in overall inference time ($1.13\times$ faster) and a 46.13\% reduction in decoder time ($1.86\times$ faster).
The speedup achieved by \method{} varies depending on the backbone size of the model.
Even for ToC3D, which features a super large backbone and high-resolution images ($1600\times 800$), \method{} still reduces overall inference time by 5.30\%, while the decoder time decreases by 49.73\% ($1.99\times$ faster)—all without compromising model performance.

NDS, which reflects the overall performance of the model, remains highly stable with \method{}, maintaining a drop within 1\%.
For instance, when applying \method{} to OPEN-ResNet50-704x256, the NDS decreases by only 0.002, demonstrating minimal impact on performance.
Interestingly, some mTPs (mean True Positives) have even decreased after applying \method{}.
For example, the mAAE of MV2D decreases from 0.1722 to 0.1717, as shown in \cref{tab:results-performance}.
For optimal and more comprehensive results with different configurations, please refer to \cref{sec:appendix:more_results:optimal_results}.

In addition to quantitative results, we provide visualizations of PETR~\cite{petr}, FocalPETR~\cite{focalPetr}, StreamPETR~\cite{streampetr}, OPEN~\cite{open}, 3DPPE~\cite{3dppe} and ToC3D~\cite{toc3d} in \cref{fig:results_visualization}.
These visualizations demonstrate that bounding boxes remain largely unchanged after applying \method{}, further validating its effectiveness.

\subsection{Comparison with ToMe}
\label{sec:experiments:comparison}

As the first method to explore zero-shot, retraining-free pruning for 3D object models, there are no directly comparable methods in the literature.
To validate the effectiveness of our approach, we benchmark \method{} against ToMe, an advanced runtime pruning technique designed to accelerate ViT-based models.
While ToMe is primarily designed for 2D vision tasks, it represents the closest existing work in terms of runtime pruning efficiency, making it a relevant baseline for comparison.
This allows us to demonstrate the advantages of \method{} in optimizing 3D models.

As shown in \cref{tab:results-comparison}, ToMe significantly degrades the performance of both StreamPETR and OPEN while offering limited improvements in inference speed.
In some cases, such as OPEN with VovNet, the computational overhead of calculating the similarity matrix outweighs the benefits, resulting in a slowdown rather than an acceleration.

For further comparisons with more models across different values of $r$ and $n$, please refer to \cref{sec:appendix:more_results:more_comparison}.

\begin{table*}[ht]
\centering
\resizebox{\textwidth}{!}{
\begin{tabular}{cc@{\hspace{5pt}}c|c@{\hspace{5pt}}c|c@{\hspace{5pt}}c|c@{\hspace{5pt}}c@{\hspace{5pt}}c@{\hspace{5pt}}c@{\hspace{5pt}}c|c@{\hspace{5pt}}c}
\toprule
  \textbf{Model} 
& \textbf{Backbone}
& \textbf{ImageSize}
& \textbf{Pruning}
& $r$
& \textbf{mAP} $\uparrow$
& \textbf{NDS} $\uparrow$
& \textbf{mATE} $\downarrow$
& \textbf{mASE} $\downarrow$
& \textbf{mAOE} $\downarrow$
& \textbf{mAVE} $\downarrow$
& \textbf{mAAE} $\downarrow$
& \textbf{Inf. Time(ms)} $\downarrow$
& \textbf{Decoder Time(ms)} $\downarrow$ \\ \midrule
\multirow{6}{*}{\begin{tabular}[c]{@{}c@{}}StreamPETR~\cite{streampetr}\\(ICCV 2023)\end{tabular}}
      & \multirow{3}{*}{ResNet50}    & \multirow{3}{*}{704x256}
      & -   & -    & 38.01\% & 0.4822 & 0.6781 & 0.2763 & 0.6401 & 0.2831 & 0.2007 & 60.52            & 31.10            \\
& & & ToMe  & 2000 & 36.12\% & 0.4651 & 0.7056 & 0.2806 & 0.6725 & 0.3011 & 0.1952 & 57.26(-5.39\%)   & 29.38(-5.53\%)   \\
& & & Ours & 2000 & 37.93\% & 0.4817 & 0.6787 & 0.2758 & 0.6390 & 0.2844 & 0.2016 & 51.68(-14.61\%)  & 24.15(-22.35\%)   \\  \cmidrule{2-14}
                            & \multirow{3}{*}{VovNet}    & \multirow{3}{*}{1600x640}
      & -   & -     & 48.89\% & 0.5732 & 0.6096 & 0.2601 & 0.3882 & 0.2603 & 0.1944 & 288.07           & 64.93            \\
& & & ToMe  & 12000 & 46.93\% & 0.5584 & 0.6301 & 0.2786 & 0.3876 & 0.2714 & 0.1951 & 277.60(-3.63\%)  & 55.63(-14.32\%)  \\ 
& & & Ours & 12000 & 48.85\% & 0.5738 & 0.6078 & 0.2603 & 0.3813 & 0.2613 & 0.1941 & 266.49(-7.47\%)  & 46.16(-28.91\%)   \\ \midrule
\multirow{9}{*}{\begin{tabular}[c]{@{}c@{}}OPEN~\cite{open}\\(ECCV 2024)\end{tabular}}
    & \multirow{3}{*}{ResNet50} & \multirow{3}{*}{704x256} 
    & -     & -    & 47.02\% & 0.5657 & 0.5676 & 0.2702 & 0.4221 & 0.2321 & 0.2019 & 77.25           & 33.63  \\
& & & ToMe  & 2000 & 45.10\% & 0.5495 & 0.5843 & 0.2719 & 0.4705 & 0.2373 & 0.1960 & 70.15(-9.19\%)  & 29.48(-12.34\%)  \\
& & & Ours  & 2000 & 46.79\% & 0.5641 & 0.5691 & 0.2711 & 0.4232 & 0.2318 & 0.2029 & 63.31(-18.05\%) & 23.94(-28.81\%)  \\ \cmidrule{2-14}
                      & \multirow{3}{*}{VovNet}  & \multirow{3}{*}{800x320} 
    & -     & -    & 52.07\% & 0.6128 & 0.5250 & 0.2566 & 0.2811 & 0.2148 & 0.1982 & 118.55          & 29.35           \\
& & & ToMe  & 3000 & 49.21\% & 0.5926 & 0.5532 & 0.2594 & 0.3002 & 0.2248 & 0.1971 & 118.63(\red{+0.07\%}) & 29.70(\red{+1.19\%})  \\
& & & Ours & 3000 & 52.12\% & 0.6130 & 0.5250 & 0.2569 & 0.2810 & 0.2148 & 0.1985 & 110.21(-7.04\%) & 22.09(-24.74\%)  \\ \cmidrule{2-14}
                      & \multirow{3}{*}{ResNet101} & \multirow{3}{*}{1408x512} 
    & -     & -    & 51.80\% & 0.6043 & 0.5314 & 0.2679 & 0.3457 & 0.2095 & 0.1922 & 196.65           & 46.39            \\
& & & ToMe  & 8000 & 50.36\% & 0.5939 & 0.5426 & 0.2676 & 0.3593 & 0.2160 & 0.1939 & 189.84(-3.46\%)  & 40.42(-12.87\%)  \\
& & & Ours & 8000 & 51.63\% & 0.6033 & 0.5336 & 0.2681 & 0.3439 & 0.2102 & 0.1927 & 184.69(-6.08\%)  & 33.76(-27.23\%)   \\
\bottomrule
\end{tabular}
}

\caption{
Compare to ToMe~\cite{tome}.
We set $n$=2 for both ToMe and \method{}, $k$ = 175 for \method{}.
}
\label{tab:results-comparison}
\vskip -0.1in
\end{table*}

\subsection{Inference Speed on OrangePi}
\label{sec:experiments:exp_edge_device}

A major challenge in deploying models for real-world applications is achieving efficient inference on edge devices.
To evaluate the impact of \method{} on edge-device performance, we deployed FocalPETR and StreamPETR on OrangePi.
As shown in \cref{tab:results_speed_on_orangepi}, \method{} improves inference efficiency.
For FocalPETR-r18-1408x512, \method{} reduces inference time by 15.47\%, making the model $1.18\times$ faster.
For StreamPETR-vov-1600x640, \method{} achieves a 16.11\% reduction in inference time, speeding up the model by $1.19\times$.

Meanwhile, quantization is also a common method for accelerating model inference.
We quantized the model to fp16 format and compared it with the non-quantized version.
The results, shown in \cref{tab:results_speed_on_orangepi}, confirm that our method can still accelerate the inference of the quantized model.

\begin{table}[t]
\centering
\resizebox{\linewidth}{!}{
\begin{tabular}{ccccc}
\toprule
\textbf{Model}
& \textbf{Quantization}
& $r$ 
& \textbf{Inf. Time (ms)} $\downarrow$
& \textbf{Time Reduce} $\uparrow$
\\ \midrule
\ml{4}{FocalPETR}  & \ml{2}{-}    & -     & 1605.28 & -               \\
                   &              & 12000 & 1406.66 & -12.37\%(1.14x) \\ \cmidrule{2-5}
                   & \ml{2}{fp16} & -     & 1600.71 & -               \\
                   &              & 12000 & 1353.09 & -15.47\%(1.18x) \\ \midrule
\ml{4}{StreamPETR} & \ml{2}{-}    & -     & 3466.29 & -               \\
                   &              & 21000 & 2946.40 & -15.00\%(1.18x) \\ \cmidrule{2-5}
                   & \ml{2}{fp16} & -     & 3446.85 & -               \\
                   &              & 21000 & 2891.40 & -16.11\%(1.19x) \\
\bottomrule
\end{tabular}
}
\caption{
Speed tested on OrangePi.
We show the results of $n=1, k=175$.
Reported FocalPETR(StreamPETR) uses ResNet18(VovNet) as backbone with an image resolution of $1408\times 512$($1600\times 640)$.
}
\label{tab:results_speed_on_orangepi}
\vskip -0.1in
\end{table}

\subsection{Ablation Study}
\label{sec:experiments:ablation}

In this section, we analyze the impact of different parameters on model performance and efficiency using StreamPETR-vov-1600x640 and OPEN-r101-1408x512 as two representative models. 

\subsubsection{Influence of Parameters $k$, $r$ and $n$}

\begin{figure}[t]
\vskip -0.05in
    \centering
    \includegraphics[width=\linewidth]{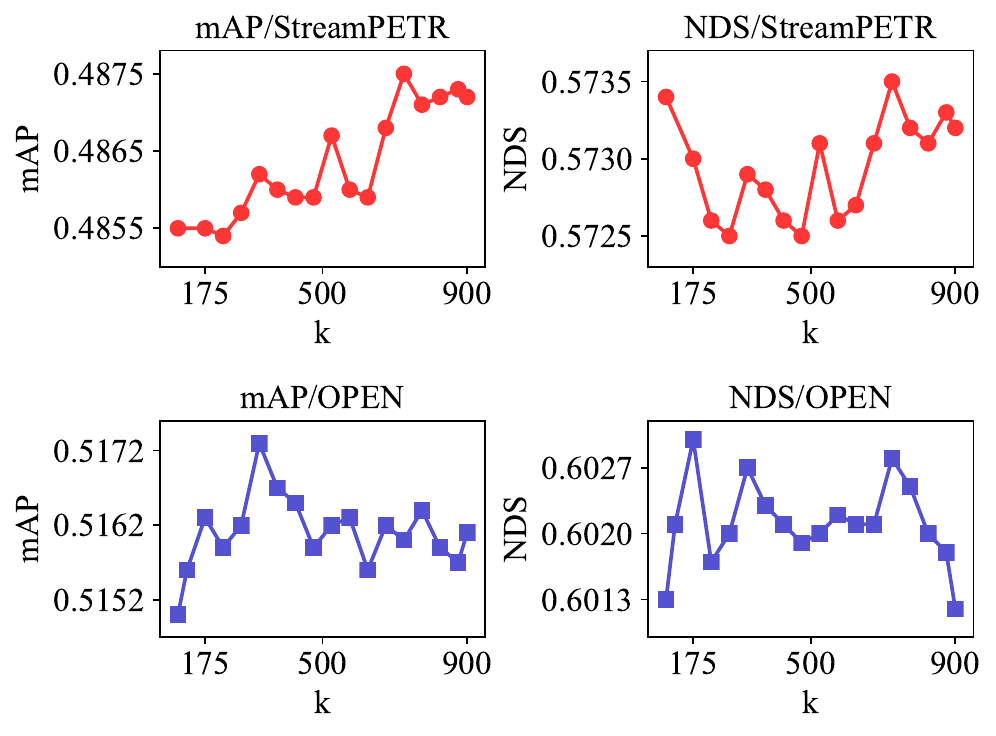}
    \vskip -0.1in
    \caption{Ablation experiments for parameter $k$, with a range from $100$ to $900$.
    We use $r=21000$ for StreamPETR, $r=10000$ for OPEN and $n=2$ for both models.
    }
    \label{fig:ablation_k}
\vskip -0.15in
\end{figure}

As illustrated in \cref{fig:ablation_k}, there is no universally optimal value of $k$ for all models.
For some models, such as StreamPETR (\cref{fig:ablation_k} Top), the mAP exhibits an approximately linear correlation with $k$.
However, for other models like OPEN (\cref{fig:ablation_k} Bottom),  variations in $k$ cause fluctuations in mAP within a certain range without a clear linear trend.
After evaluating all experimental results, we determine that a reference value of $k=175$ offers a reasonable balance between different models.
$k$ affects little to the inference time.
The analysis can be found in \cref{sec:appendix:analysis}.

\begin{figure}[t]
    \centering
    \includegraphics[width=\linewidth]{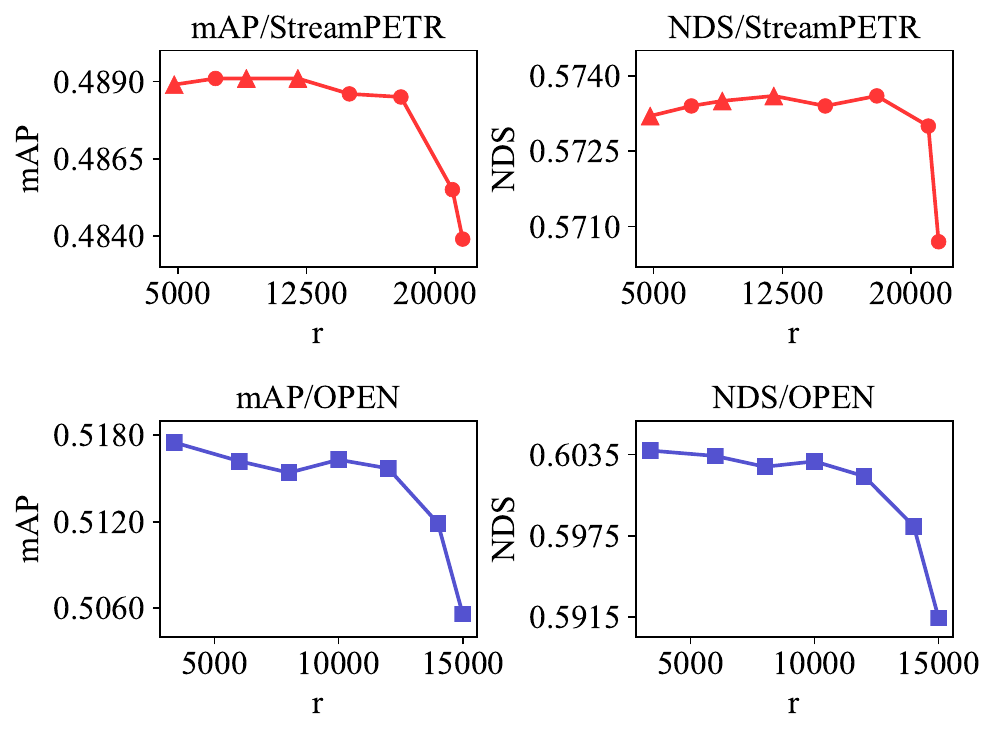}
    \vskip -0.1in
    \caption{
    The mAP and NDS exhibit a negative correlation with the increase of \( r \).
    When \( r \) exceeds a model-dependent threshold, the rate of performance degradation will accelerate sharply.
    }
    \label{fig:ablation_r}
    \vskip -0.2in
\end{figure}
\begin{figure}[t]
    \centering
    \includegraphics[width=\linewidth]{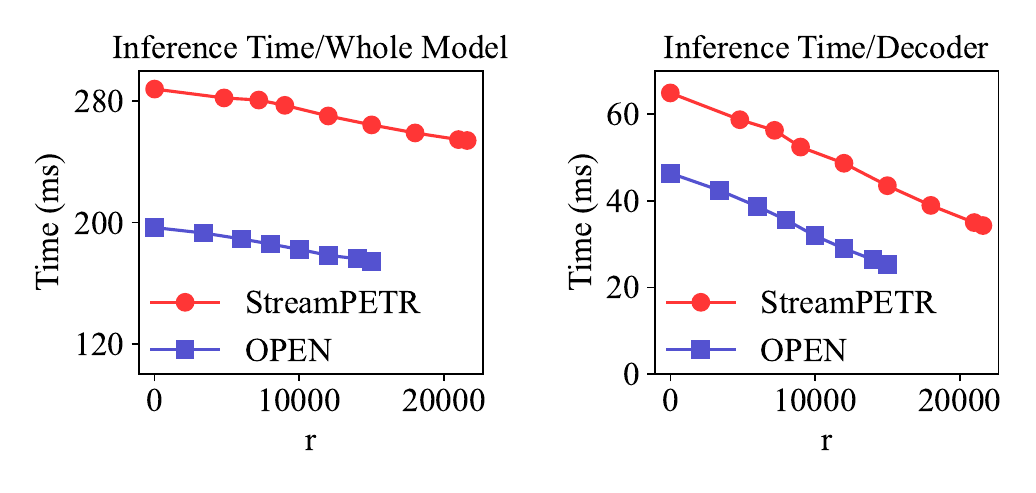}
    \vskip -0.2in
    \caption{
    The inference time of the whole model and decoder time decrease linearly with the increase of \( r \).
    }
    \label{fig:ablation_speed_r}
    \vskip -0.05in
\end{figure}

The parameter $r$ represents the number of pruned keys.
Intuitively, increasing \( r \) is expected to lead to performance degradation; however, the trend is not strictly monotonic in our experiments.
For instance, for the StreamPETR, the mAP and the NDS at $r$ = 9000 are equivalent to that at $r$ = 12000 and even higher than that at $r$ = 4800, as indicated by the points with triangles in \cref{fig:ablation_r}.
Overall, as $r$ increases, model performance declines linearly.
However, beyond a model-dependent threshold (about 80\%-90\% out of the $N_k$), the degradation rate accelerates significantly, as depicted in \cref{fig:ablation_r}.
For time reduction, there is an obvious trade-off between performance and inference time while $r$ increases, as shown in \cref{fig:ablation_r} and \cref{fig:ablation_speed_r}.

\begin{table}[t]
\resizebox{\linewidth}{!}{
\begin{tabular}{cccccc}
\toprule
  \textbf{Model}  & $n$ & \textbf{mAP $\uparrow$} & \textbf{NDS $\uparrow$} & \textbf{Inf. Time(ms)} & \textbf{Decoder Time(ms)} \\ \midrule
\multirow{6}{*}{StreamPETR}
  & - & 48.89\% & 0.5732 & 288.07           & 64.93           \\
  & 5 & 48.92\% & 0.5744 & 267.40( -7.18\%) & 46.59(-28.25\%) \\
  & 4 & 48.87\% & 0.5745 & 263.60( -8.49\%) & 44.25(-31.85\%) \\
  & 3 & 48.21\% & 0.5702 & 259.08(-10.06\%) & 39.08(-39.81\%) \\
  & 2 & 48.55\% & 0.5730 & 254.73(-11.57\%) & 34.98(-46.13\%) \\
  & 1 & 47.84\% & 0.5695 & 251.24(-12.79\%) & 31.61(-51.32\%) \\ \midrule
\multirow{6}{*}{OPEN}
  & - & 51.80\% & 0.6043 & 196.65          & 46.39           \\
  & 5 & 51.73\% & 0.6037 & 190.07(-3.35\%) & 39.51(-14.83\%) \\
  & 4 & 51.72\% & 0.6040 & 185.94(-5.45\%) & 36.29(-21.77\%) \\
  & 3 & 51.61\% & 0.6021 & 183.77(-6.55\%) & 33.59(-27.59\%) \\
  & 2 & 51.63\% & 0.6030 & 182.24(-7.33\%) & 31.92(-31.19\%) \\
  & 1 & 51.65\% & 0.6027 & 179.55(-8.70\%) & 30.01(-35.31\%) \\
\bottomrule
\end{tabular}
}
\caption{Ablation experiments for parameter $n$ with $r=21000~(10000)$ and $k=175$ for StreamPETR (OPEN).}
\label{tab:result_abl_n}
\vskip -0.15in
\end{table}

The parameter \( n \) represents the number of \method{} layers inserted into the model. 
A smaller \( n \) (with a minimum value of 1) means fewer \method{} layers, requiring each layer to prune a larger number of keys. 
Intuitively, there is a trade-off between \( n \), inference speed, and model performance: a lower \( n \) results in faster inference but can lead to greater performance degradation.
As shown in \cref{tab:result_abl_n}, this argument is supported.
However, similar to \( r \), model performance does not decrease monotonically with increasing \( n \).  Instead, it fluctuates within a certain range, suggesting that the impact of \( n \) is model-dependent.

\subsubsection{Impact of Classification Scores}

\noindent\bd{Remove Classification Scores}.
To further evaluate the rationale behind the pruning criterion, we investigate the role of the classification score in the key selection process.
Specifically, we examine whether incorporating classification scores is necessary for determining key importance.
As shown in \cref{equ:criterion_remove_cls_scores}, we remove the classification score and instead compute key importance solely based on the column-wise sum of averaged attention weights:
\begin{equation}
        S_{j} = \sum\limits_{i=1}^{N_q}\left(
            \cfrac{1}{N_h}\sum\limits_{i=1}^{N_q} A^h_{i,j}
        \right)
    \label{equ:criterion_remove_cls_scores}
\end{equation}

\begin{figure}[t]
    \centering
    \includegraphics[width=\linewidth]{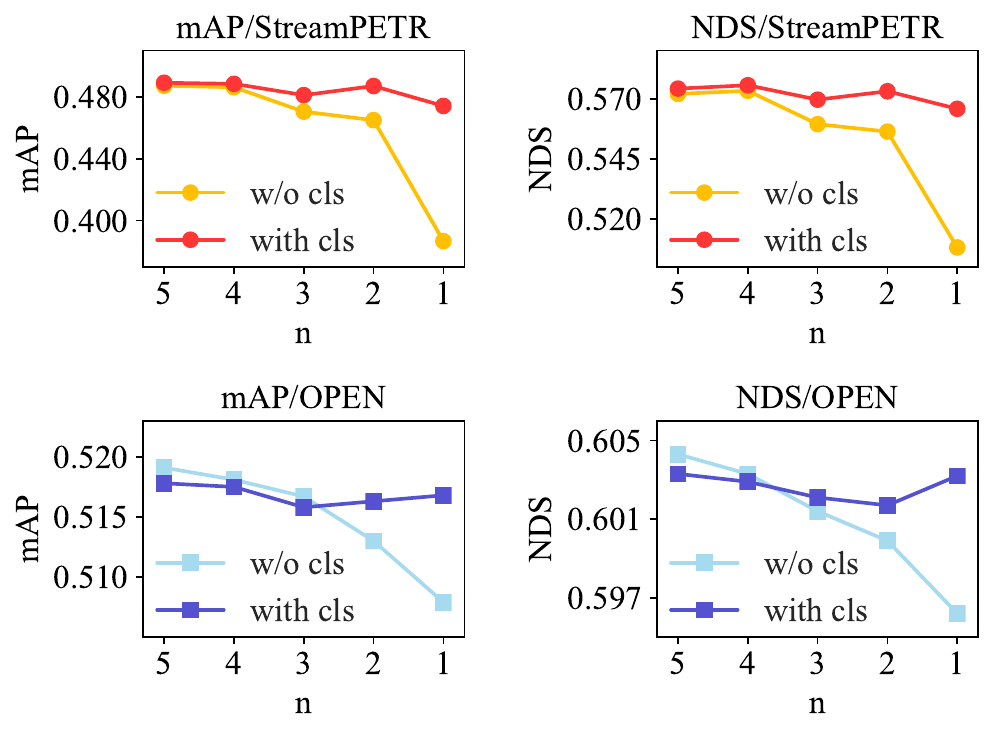}
    \vskip -0.1in
    \caption{mAP and NDS for \method{} with and w/o classification scores.
    We use $r$=21000(10000) for StreamPETR(OPEN).
    }\label{fig:ablation_cls}
    \vskip -0.05in
\end{figure}

Since classification scores are excluded, we use \( k = 900 \) in \method{} for comparison. 
As illustrated in \cref{fig:ablation_cls}, omitting the classification score leads to a significant drop in model performance as $n$ decreases. 
This result highlights the importance of incorporating classification scores in maintaining the model's stability.

\begin{table}[t]
\centering
\resizebox{\linewidth}{!}{
\begin{tabular}{c|c@{\hspace{5pt}}c|c|c@{\hspace{5pt}}c@{\hspace{5pt}}c@{\hspace{5pt}}c@{\hspace{5pt}}c}
\toprule
\textbf{Model}
& \bd{Select}
& \bd{mAP} $\uparrow$
& \textbf{NDS} $\uparrow$
& \bd{mATE} $\downarrow$
& \bd{mASE} $\downarrow$
& \bd{mAOE} $\downarrow$
& \bd{mAVE} $\downarrow$
& \bd{mAAE} $\downarrow$ \\ \midrule
\ml{4}{StreamPETR} & -    & 48.89\% & 0.5732 & 0.6096 & 0.2601 & 0.3882 & 0.2603 & 0.1944 \\
                   & max  & 48.55\% & 0.5730 & 0.6033 & 0.2626 & 0.3771 & 0.2611 & 0.1941 \\
                   & mean & 44.12\% & 0.5537 & 0.6452 & 0.2721 & 0.3799 & 0.2777 & 0.1944 \\
                   & min  & 34.88\% & 0.4760 & 0.7199 & 0.2882 & 0.4694 & 0.3082 & 0.1977 \\ \midrule
\ml{4}{OPEN}       & -    & 51.80\% & 0.6043 & 0.5314 & 0.2679 & 0.3457 & 0.2095 & 0.1922 \\
                   & max  & 51.57\% & 0.6019 & 0.5368 & 0.2726 & 0.3493 & 0.2102 & 0.1905 \\
                   & mean & 49.57\% & 0.5839 & 0.5763 & 0.2795 & 0.3688 & 0.2210 & 0.1938 \\
                   & min  & 47.65\% & 0.5703 & 0.6020 & 0.2837 & 0.3728 & 0.2299 & 0.1906 \\
\bottomrule
\end{tabular}
}
\caption{Results of ablation experiments of the selection of classification scores, with a configuration of $n$=2, $k$=175 and $r$=21000(12000) for StreamPETR(OPEN).
}
\label{tab:results_abl_cls_select}
\vskip -0.15in
\end{table}

\noindent\bd{The Selection of Classification Scores}.
The first step of \method{} involves selecting the maximum value from each classification vector, resulting in $\hat C\in\mathbb R^{N_q}$, as described in \cref{eq:expand_cls_scores}.
To validate this point, we modify \cref{eq:expand_cls_scores} by replacing the maximum selection with the average and minimum values of each classification vector.
Results, presented in  \cref{tab:results_abl_cls_select}, confirm that selecting the maximum classification score yields superior performance, supporting the design choice in \method{}.

For additional ablation studies and further exploration (Apply \method{}  to prune queries,  evaluate it on fully converged models, integrate it into training,   and extend it to 2D  models), please refer to \cref{sec:appendix:more_results:more_ablation_exps} and \cref{sec:appendix:further_exploration}.
\section{Conclusion}
\label{sec:conclusion}

In this paper, we propose \method{}, a runtime pruning method that removes redundant keys from transformer layers in 3D detectors.
\Method{} requires no retraining or data calibration, making it a simple plug-and-play solution.
Extensive experiments on multiple of state-of-the-art detectors demonstrate that our method effectively accelerates inference speed while preserving model performance, highlighting its practicality for real-world applications.
We hope this work encourages further research into zero-shot pruning techniques for 3D object detection.

\clearpage

{\small
\bibliographystyle{ieeenat_fullname}
\bibliography{main}

\begin{thebibliography}{41}
\providecommand{\natexlab}[1]{#1}
\providecommand{\url}[1]{\texttt{#1}}
\expandafter\ifx\csname urlstyle\endcsname\relax
  \providecommand{\doi}[1]{doi: #1}\else
  \providecommand{\doi}{doi: \begingroup \urlstyle{rm}\Url}\fi

\bibitem[Bolya et~al.(2023)Bolya, Fu, Dai, Zhang, Feichtenhofer, and Hoffman]{tome}
Daniel Bolya, Cheng-Yang Fu, Xiaoliang Dai, Peizhao Zhang, Christoph Feichtenhofer, and Judy Hoffman.
\newblock Token merging: Your vit but faster.
\newblock In \emph{International Conference on Learning Representations}, 2023.

\bibitem[Caesar et~al.(2020)Caesar, Bankiti, Lang, Vora, Liong, Xu, Krishnan, Pan, Baldan, and Beijbom]{nuscenes}
Holger Caesar, Varun Bankiti, Alex~H. Lang, Sourabh Vora, Venice~Erin Liong, Qiang Xu, Anush Krishnan, Yu Pan, Giancarlo Baldan, and Oscar Beijbom.
\newblock nuscenes: A multimodal dataset for autonomous driving.
\newblock In \emph{Proceedings of the IEEE/CVF Conference on Computer Vision and Pattern Recognition (CVPR)}, 2020.

\bibitem[Carion et~al.(2020)Carion, Massa, Synnaeve, Usunier, Kirillov, and Zagoruyko]{detr}
Nicolas Carion, Francisco Massa, Gabriel Synnaeve, Nicolas Usunier, Alexander Kirillov, and Sergey Zagoruyko.
\newblock End-to-end object detection with transformers.
\newblock In \emph{European Conference on Computer Vision}, pages 213--229. Springer, 2020.

\bibitem[Chang et~al.(2023)Chang, Wang, Xu, Chen, Yang, and Zhao]{detrdistill}
Jiahao Chang, Shuo Wang, Hai-Ming Xu, Zehui Chen, Chenhongyi Yang, and Feng Zhao.
\newblock Detrdistill: A universal knowledge distillation framework for detr-families.
\newblock In \emph{Proceedings of the IEEE/CVF International Conference on Computer Vision}, pages 6898--6908, 2023.

\bibitem[Chen et~al.(2024)Chen, Ma, Qiao, and Wang]{mbev}
Siran Chen, Yue Ma, Yu Qiao, and Yali Wang.
\newblock M-bev: Masked bev perception for robust autonomous driving.
\newblock In \emph{Proceedings of the AAAI Conference on Artificial Intelligence}, 2024.

\bibitem[Chen et~al.(2021)Chen, Cheng, Gan, Yuan, Zhang, and Wang]{svite}
Tianlong Chen, Yu Cheng, Zhe Gan, Lu Yuan, Lei Zhang, and Zhangyang Wang.
\newblock Chasing sparsity in vision transformers: An end-to-end exploration.
\newblock \emph{Advances in Neural Information Processing Systems}, 34:\penalty0 19974--19988, 2021.

\bibitem[Dosovitskiy(2020)]{vit}
Alexey Dosovitskiy.
\newblock An image is worth 16x16 words: Transformers for image recognition at scale.
\newblock \emph{arXiv preprint arXiv:2010.11929}, 2020.

\bibitem[Fan et~al.(2019)Fan, Grave, and Joulin]{layerdrop}
Angela Fan, Edouard Grave, and Armand Joulin.
\newblock Reducing transformer depth on demand with structured dropout.
\newblock \emph{arXiv preprint arXiv:1909.11556}, 2019.

\bibitem[Fang et~al.(2024)Fang, Sun, Wang, Huang, Wang, and Cao]{eva02}
Yuxin Fang, Quan Sun, Xinggang Wang, Tiejun Huang, Xinlong Wang, and Yue Cao.
\newblock Eva-02: A visual representation for neon genesis.
\newblock \emph{Image and Vision Computing}, 149:\penalty0 105171, 2024.

\bibitem[Fayyaz et~al.(2022)Fayyaz, Koohpayegani, Jafari, Sengupta, Joze, Sommerlade, Pirsiavash, and Gall]{ats}
Mohsen Fayyaz, Soroush~Abbasi Koohpayegani, Farnoush~Rezaei Jafari, Sunando Sengupta, Hamid Reza~Vaezi Joze, Eric Sommerlade, Hamed Pirsiavash, and J{\"u}rgen Gall.
\newblock Adaptive token sampling for efficient vision transformers.
\newblock In \emph{European Conference on Computer Vision}, pages 396--414. Springer, 2022.

\bibitem[Frankle and Carbin(2018)]{lth}
Jonathan Frankle and Michael Carbin.
\newblock The lottery ticket hypothesis: Finding sparse, trainable neural networks.
\newblock \emph{arXiv preprint arXiv:1803.03635}, 2018.

\bibitem[He et~al.(2016)He, Zhang, Ren, and Sun]{resnet}
Kaiming He, Xiangyu Zhang, Shaoqing Ren, and Jian Sun.
\newblock Deep residual learning for image recognition.
\newblock In \emph{Proceedings of the IEEE conference on computer vision and pattern recognition}, pages 770--778, 2016.

\bibitem[Hinton et~al.(2015)Hinton, Vinyals, and Dean]{distilling_learning}
Geoffrey Hinton, Oriol Vinyals, and Jeff Dean.
\newblock Distilling the knowledge in a neural network.
\newblock \emph{arXiv preprint arXiv:1503.02531}, 2015.

\bibitem[Hou et~al.(2024)Hou, Wang, Ye, Liu, Tan, Ding, Wang, and Bai]{open}
Jinghua Hou, Tong Wang, Xiaoqing Ye, Zhe Liu, Xiao Tan, Errui Ding, Jingdong Wang, and Xiang Bai.
\newblock Open: Object-wise position embedding for multi-view 3d object detection.
\newblock In \emph{European Conference on Computer Vision}, 2024.

\bibitem[Jiang et~al.(2024)Jiang, Li, Liu, Wang, Jia, Wang, Han, and Zhang]{far3d}
Xiaohui Jiang, Shuailin Li, Yingfei Liu, Shihao Wang, Fan Jia, Tiancai Wang, Lijin Han, and Xiangyu Zhang.
\newblock Far3d: Expanding the horizon for surround-view 3d object detection.
\newblock In \emph{Proceedings of the AAAI Conference on Artificial Intelligence}, pages 2561--2569, 2024.

\bibitem[Khaki and Plataniotis(2024)]{optin}
Samir Khaki and Konstantinos~N Plataniotis.
\newblock The need for speed: Pruning transformers with one recipe.
\newblock In \emph{The Twelfth International Conference on Learning Representations}, 2024.

\bibitem[Kwon et~al.(2022)Kwon, Kim, Mahoney, Hassoun, Keutzer, and Gholami]{fast_post_traing_pruning}
Woosuk Kwon, Sehoon Kim, Michael~W Mahoney, Joseph Hassoun, Kurt Keutzer, and Amir Gholami.
\newblock A fast post-training pruning framework for transformers.
\newblock \emph{Advances in Neural Information Processing Systems}, 35:\penalty0 24101--24116, 2022.

\bibitem[Lee and Park(2020)]{vovnetv2}
Youngwan Lee and Jongyoul Park.
\newblock Centermask: Real-time anchor-free instance segmentation.
\newblock In \emph{Proceedings of the IEEE/CVF conference on computer vision and pattern recognition}, pages 13906--13915, 2020.

\bibitem[Li et~al.(2023)Li, Ge, Yu, Yang, Wang, Shi, Sun, and Li]{bevdepth}
Yinhao Li, Zheng Ge, Guanyi Yu, Jinrong Yang, Zengran Wang, Yukang Shi, Jianjian Sun, and Zeming Li.
\newblock Bevdepth: Acquisition of reliable depth for multi-view 3d object detection.
\newblock In \emph{Proceedings of the AAAI Conference on Artificial Intelligence}, pages 1477--1485, 2023.

\bibitem[Li et~al.(2022)Li, Wang, Li, Xie, Sima, Lu, Qiao, and Dai]{bevformer}
Zhiqi Li, Wenhai Wang, Hongyang Li, Enze Xie, Chonghao Sima, Tong Lu, Yu Qiao, and Jifeng Dai.
\newblock Bevformer: Learning bird’s-eye-view representation from multi-camera images via spatiotemporal transformers.
\newblock In \emph{European Conference on Computer Vision}, pages 1--18. Springer, 2022.

\bibitem[Liang et~al.(2023)Liang, Zuo, Zhang, He, Chen, and Zhao]{task_aware_distill}
Chen Liang, Simiao Zuo, Qingru Zhang, Pengcheng He, Weizhu Chen, and Tuo Zhao.
\newblock Less is more: Task-aware layer-wise distillation for language model compression.
\newblock In \emph{International Conference on Machine Learning}, pages 20852--20867. PMLR, 2023.

\bibitem[Liu et~al.(2024{\natexlab{a}})Liu, Huang, Zhang, Yao, Zhang, Wan, Ye, and Zhou]{raydn}
Feng Liu, Tengteng Huang, Qianjing Zhang, Haotian Yao, Chi Zhang, Fang Wan, Qixiang Ye, and Yanzhao Zhou.
\newblock Ray denoising: Depth-aware hard negative sampling for multi-view 3d object detection.
\newblock In \emph{European Conference on Computer Vision}. Springer, 2024{\natexlab{a}}.

\bibitem[Liu et~al.(2023{\natexlab{a}})Liu, Teng, Lu, Wang, and Wang]{sparsebev}
Haisong Liu, Yao Teng, Tao Lu, Haiguang Wang, and Limin Wang.
\newblock Sparsebev: High-performance sparse 3d object detection from multi-camera videos.
\newblock In \emph{Proceedings of the IEEE/CVF International Conference on Computer Vision}, pages 18580--18590, 2023{\natexlab{a}}.

\bibitem[Liu et~al.(2022)Liu, Wang, Zhang, and Sun]{petr}
Yingfei Liu, Tiancai Wang, Xiangyu Zhang, and Jian Sun.
\newblock Petr: Position embedding transformation for multi-view 3d object detection.
\newblock In \emph{European Conference on Computer Vision}, pages 531--548. Springer, 2022.

\bibitem[Liu et~al.(2023{\natexlab{b}})Liu, Yan, Jia, Li, Gao, Wang, and Zhang]{petrv2}
Yingfei Liu, Junjie Yan, Fan Jia, Shuailin Li, Aqi Gao, Tiancai Wang, and Xiangyu Zhang.
\newblock Petrv2: A unified framework for 3d perception from multi-camera images.
\newblock In \emph{Proceedings of the IEEE/CVF International Conference on Computer Vision (ICCV)}, pages 3262--3272, 2023{\natexlab{b}}.

\bibitem[Liu et~al.(2024{\natexlab{b}})Liu, Gehrig, Messikommer, Cannici, and Scaramuzza]{svit}
Yifei Liu, Mathias Gehrig, Nico Messikommer, Marco Cannici, and Davide Scaramuzza.
\newblock Revisiting token pruning for object detection and instance segmentation.
\newblock In \emph{Proceedings of the IEEE/CVF Winter Conference on Applications of Computer Vision}, pages 2658--2668, 2024{\natexlab{b}}.

\bibitem[Michel et~al.(2019)Michel, Levy, and Neubig]{sixteen}
Paul Michel, Omer Levy, and Graham Neubig.
\newblock Are sixteen heads really better than one?
\newblock \emph{Advances in neural information processing systems}, 32, 2019.

\bibitem[Shu et~al.(2023)Shu, Deng, Yu, and Liu]{3dppe}
Changyong Shu, Jiajun Deng, Fisher Yu, and Yifan Liu.
\newblock 3dppe: 3d point positional encoding for multi-camera 3d object detection transformers.
\newblock In \emph{Proceedings of the IEEE/CVF International Conference on Computer Vision}, 2023.

\bibitem[Touvron et~al.(2021)Touvron, Cord, Douze, Massa, Sablayrolles, and Jegou]{deit}
Hugo Touvron, Matthieu Cord, Matthijs Douze, Francisco Massa, Alexandre Sablayrolles, and Herve Jegou.
\newblock Training data-efficient image transformers \& distillation through attention.
\newblock In \emph{International Conference on Machine Learning}, pages 10347--10357, 2021.

\bibitem[Vaswani et~al.(2017)Vaswani, Shazeer, Parmar, Uszkoreit, Jones, Gomez, Kaiser, and Polosukhin]{transformer}
Ashish Vaswani, Noam Shazeer, Niki Parmar, Jakob Uszkoreit, Llion Jones, Aidan~N Gomez, {\L}ukasz Kaiser, and Illia Polosukhin.
\newblock Attention is all you need.
\newblock \emph{Advances in neural information processing systems}, 30, 2017.

\bibitem[Wang et~al.(2024)Wang, Dedhia, and Jha]{zero_tprune}
Hongjie Wang, Bhishma Dedhia, and Niraj~K Jha.
\newblock Zero-tprune: Zero-shot token pruning through leveraging of the attention graph in pre-trained transformers.
\newblock In \emph{Proceedings of the IEEE/CVF Conference on Computer Vision and Pattern Recognition}, pages 16070--16079, 2024.

\bibitem[Wang et~al.(2023{\natexlab{a}})Wang, Jiang, and Li]{focalPetr}
Shihao Wang, Xiaohui Jiang, and Ying Li.
\newblock Focal-petr: Embracing foreground for efficient multi-camera 3d object detection.
\newblock \emph{IEEE Transactions on Intelligent Vehicles}, 2023{\natexlab{a}}.

\bibitem[Wang et~al.(2023{\natexlab{b}})Wang, Liu, Wang, Li, and Zhang]{streampetr}
Shihao Wang, Yingfei Liu, Tiancai Wang, Ying Li, and Xiangyu Zhang.
\newblock Exploring object-centric temporal modeling for efficient multi-view 3d object detection.
\newblock In \emph{Proceedings of the IEEE/CVF International Conference on Computer Vision}, pages 3621--3631, 2023{\natexlab{b}}.

\bibitem[Wang et~al.(2022)Wang, Guizilini, Zhang, Wang, Zhao, and Solomon]{detr3d}
Yue Wang, Vitor~Campagnolo Guizilini, Tianyuan Zhang, Yilun Wang, Hang Zhao, and Justin Solomon.
\newblock Detr3d: 3d object detection from multi-view images via 3d-to-2d queries.
\newblock In \emph{Conference on Robot Learning}, pages 180--191. PMLR, 2022.

\bibitem[Wang et~al.(2023{\natexlab{c}})Wang, Huang, Fu, Wang, and Liu]{mv2d}
Zitian Wang, Zehao Huang, Jiahui Fu, Naiyan Wang, and Si Liu.
\newblock Object as query: Lifting any 2d object detector to 3d detection.
\newblock In \emph{Proceedings of the IEEE/CVF International Conference on Computer Vision}, pages 3791--3800, 2023{\natexlab{c}}.

\bibitem[Xu et~al.(2024)Xu, Pang, Qiu, Wu, Bai, Mei, and Xue]{gpq}
Lizhen Xu, Shanmin Pang, Wenzhao Qiu, Zehao Wu, Xiuxiu Bai, Kuizhi Mei, and Jianru Xue.
\newblock Redundant queries in detr-based 3d detection methods: Unnecessary and prunable, 2024.

\bibitem[Yang et~al.(2023)Yang, Chen, Tian, Tao, Zhu, Zhang, Huang, Li, Qiao, Lu, Zhou, and Dai]{bevformerv2}
Chenyu Yang, Yuntao Chen, Hao Tian, Chenxin Tao, Xizhou Zhu, Zhaoxiang Zhang, Gao Huang, Hongyang Li, Yu Qiao, Lewei Lu, Jie Zhou, and Jifeng Dai.
\newblock Bevformer v2: Adapting modern image backbones to bird's-eye-view recognition via perspective supervision.
\newblock In \emph{2023 IEEE/CVF Conference on Computer Vision and Pattern Recognition (CVPR)}, pages 17830--17839, 2023.

\bibitem[Yu et~al.(2022)Yu, Huang, Wang, Cheng, Chu, and Cui]{wdpruning}
Fang Yu, Kun Huang, Meng Wang, Yuan Cheng, Wei Chu, and Li Cui.
\newblock Width \& depth pruning for vision transformers.
\newblock In \emph{Proceedings of the AAAI Conference on Artificial Intelligence}, pages 3143--3151, 2022.

\bibitem[Yu and Xiang(2023)]{xpruner}
Lu Yu and Wei Xiang.
\newblock X-pruner: explainable pruning for vision transformers.
\newblock In \emph{Proceedings of the IEEE/CVF conference on computer vision and pattern recognition}, pages 24355--24363, 2023.

\bibitem[Zhang et~al.(2024)Zhang, Liang, Tan, Ye, Zhang, Wang, and Bai]{toc3d}
Dingyuan Zhang, Dingkang Liang, Zichang Tan, Xiaoqing Ye, Cheng Zhang, Jingdong Wang, and Xiang Bai.
\newblock Make your vit-based multi-view 3d detectors faster via token compression.
\newblock In \emph{European Conference on Computer Vision}, 2024.

\bibitem[Zhu et~al.(2021)Zhu, Su, Lu, Li, Wang, and Dai]{deformabledetr}
Xizhou Zhu, Weijie Su, Lewei Lu, Bin Li, Xiaogang Wang, and Jifeng Dai.
\newblock Deformable detr: Deformable transformers for end-to-end object detection.
\newblock In \emph{International Conference on Learning Representations}, 2021.

\end{thebibliography}
}

\clearpage
\setcounter{page}{1}
\maketitlesupplementary

\appendix
\section{Query-based 3D Detectors}
\label{sec:appendix:3d_detectors}

\subsection{Overall Architecture}
\label{sec:appendix:3d_detectors:overall_arch}

\begin{figure}[t]
    \centering
    \includegraphics[width=\linewidth]{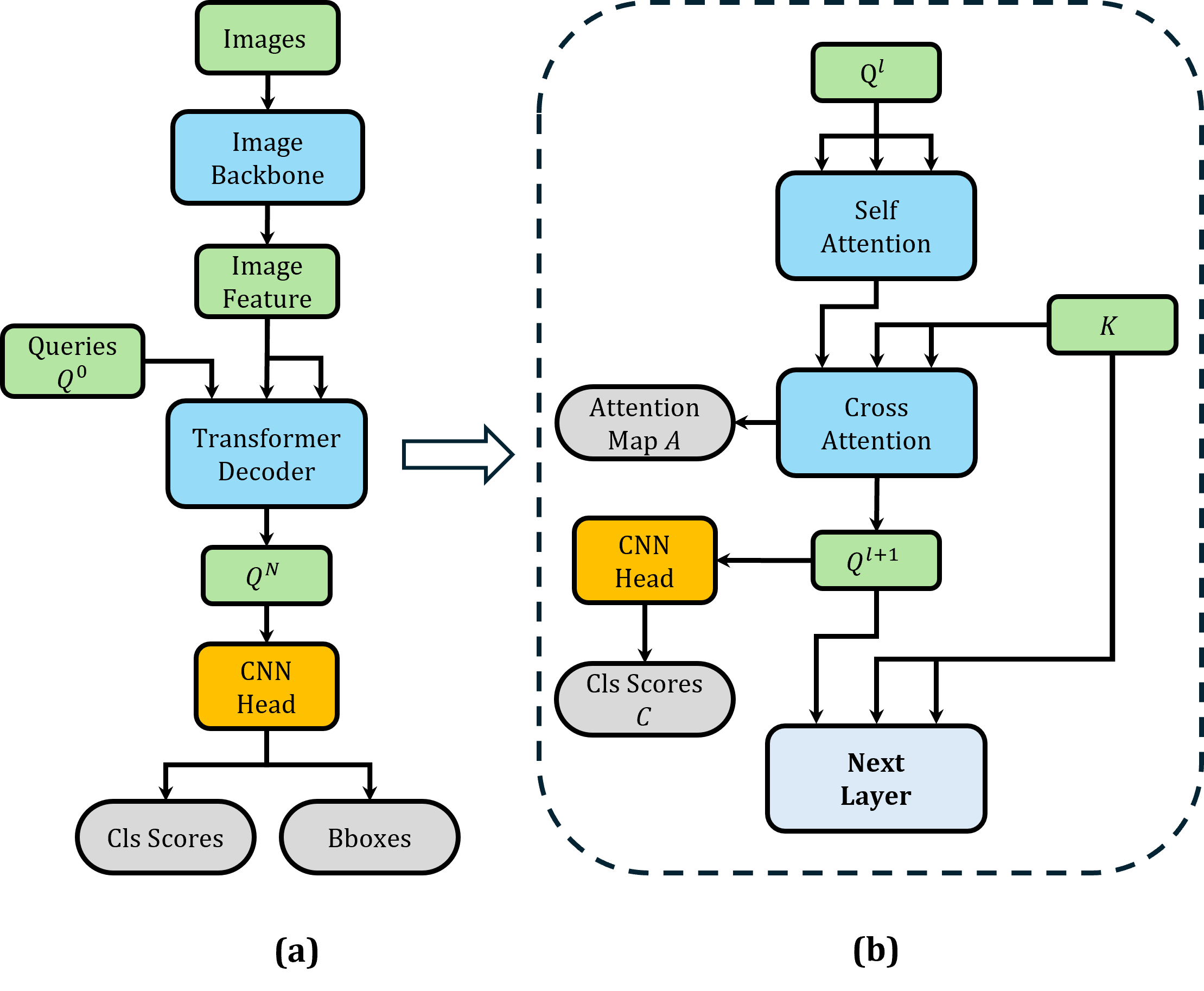}
    \caption{
      \textbf{(a)} Overall architecture of query-based 3D detectors.
      All dense models share the same overall structure: an image backbone followed by a transformer decoder.
      The transformer decoder consists of multiple stacked transformer layers.
      \textbf{(b)} The transformer layer used in the transformer decoder in (a).
      Each layer has a self-attention module and a cross-attention module.
      A CNN head can be used following the cross-attention module, which inputs updated queries and outputs the classification scores.
      We use attention map $A\in\mathbb R^{N_q\times N_k}$ and classification scores $C\in\mathbb R^{N_q}$
      to calculate each key's importance score.
    }
    \label{fig:3d_detector}
\vskip -0.1in
\end{figure}

DETR-based methods have become the mainstream approach for 3D object detection.  
As shown in \cref{fig:3d_detector} (a), these methods typically take multi-view images as input and use an image backbone to extract image features \( F \).
Subsequently, \( F \) is fed into a transformer decoder along with predefined queries \( Q \) for interaction.  
Represented by PETR, \bd{dense methods} adopt global attention, allowing \( Q \) to interact with \( F \) globally.
In contrast, \bd{sparse methods} such as DETR3D and Far3D employ deformable attention, selecting only a subset of \( F \) for interaction with \( Q \). 
The output of the Transformer Decoder maintains the same shape as the predefined queries \( Q \).
It is passed to classification branches and regression branches to obtain classification scores and bounding boxes \( B \).
The bounding boxes \( B \) not only contain object location information but also include object size, orientation, velocity, and additional attributes (e.g., whether a pedestrian is standing, walking, or sitting).  
During training, DETR-based methods use a bipartite matching strategy to associate predictions with ground truth and compute classification and localization losses.
Since ground truth is unavailable during inference, the model selects the final predictions based on classification scores.

The Transformer decoder used in 3D detectors has a structure of self-attention followed by cross-attention.
The self-attention module of the first layer takes pre-defined query $Q$ as input, and its output, together with image features, will be fed into the following cross-attention module.
The output of the cross-attention module will be taken as input by the next layer's self-attention module.
Our method uses the attention map generated by the cross-attention modules.

\begin{table*}[ht]
\resizebox{\textwidth}{!}{
\begin{tabular}{ccc|cc|ccccc|cc}
\toprule
  \bd{Model}
& \bd{Backbone}
& \bd{ImageSize}
& \bd{mAP} $\uparrow$
& \bd{NDS} $\uparrow$
& \bd{mATE} $\downarrow$
& \bd{mASE} $\downarrow$
& \bd{mAOE} $\downarrow$
& \bd{mAVE} $\downarrow$
& \bd{mAAE} $\downarrow$
& \bd{Mem. (MiB)} $\downarrow$
& \bd{Inf. Time (ms)} $\downarrow$ \\ \midrule
\ml{2}{SparseBEV} & ResNet50  & 704x256   & 45.45\% & 0.5559 & 0.5984 & 0.2706 & 0.4124 & 0.2435 & 0.1865 & 5640       & 77.72   \\
                  & ResNet101 & 1408x512  & 50.12\% & 0.5920 & 0.5621 & 0.2648 & 0.3211 & 0.2427 & 0.1947 & 15386      & 193.25  \\ \midrule
\ml{2}{OPEN}      & ResNet50  & 704x256   & 47.02\% & 0.5657 & 0.5676 & 0.2702 & 0.4221 & 0.2321 & 0.2019 & 4062       & 77.25   \\
                  & ResNet101 & 1408x512  & 51.80\% & 0.6043 & 0.5314 & 0.2679 & 0.3457 & 0.2095 & 0.1922 & 10696      & 196.55  \\
\bottomrule
\end{tabular}}
\caption{
Comparison of SparseBEV and OPEN.
}
\label{tab:comparison_of_sparse_dense_models}
\end{table*}

\subsection{Advantages of Dense Methods over Sparse Methods}
\label{sec:appendix:3d_detectors:advantages_over_sparse_methods}

Since the introduction of DETR, DETR-based methods have gradually gained prominence in various vision tasks, including image classification, 2D object detection, 3D object detection, semantic segmentation, and object tracking.  
However, DETR-based methods also have several drawbacks, such as high computational cost and slow inference speed. One major issue is their slow convergence rate. DETR requires training for several hundred epochs on the COCO dataset before convergence.  
To address this, DeformableDETR was proposed.
By utilizing reference points for local feature sampling, DeformableDETR significantly speeds up convergence, requiring only 50 epochs on the COCO dataset to achieve the same performance as DETR trained for 500 epochs.
In the 3D object detection domain, sparse methods have been introduced to accelerate model convergence while also reducing the computational burden of the decoder to some extent.

However, our comparison reveals that the overall model does not achieve a significant increase in inference speed, nor does it reduce memory consumption.
In some cases, sacrificing performance necessitates additional compensatory mechanisms, introducing extra parameters that increase memory usage.  

We selected state-of-the-art dense and sparse methods, OPEN and SparseBEV, respectively, for comparison.
From the comparison in \cref{tab:comparison_of_sparse_dense_models}, it can be observed that when using the same backbone and image resolution, SparseBEV performs worse than OPEN on the nuScenes validation set.
In terms of memory usage, during inference, SparseBEV consumes more memory than OPEN when using the same backbone and image resolution.  
Regarding inference speed, SparseBEV does not show a significant advantage over OPEN.

In summary, while current dense methods converge more slowly than sparse methods, they offer advantages in performance and memory usage, without showing a clear disadvantage in inference speed.  
Due to their strong overall advantages, new dense methods continue to emerge.
This further highlights the importance of our work.

\section{Analysis of Computation Cost of Importance Scores}
\label{sec:appendix:analysis}


\bd{The FLOPs of Matrix Multiplication}.
Assume there are two matrices \(A\in\mathbb R^{N\times C}\), \(B\in\mathbb R^{M\times C}\).
When computing \(AB^T\in\mathbb R^{N\times M}\), each row \(A_i\in\mathbb R^{C}\) of \(A\) is multiplied by each column \(B_j\in\mathbb R^C\) of \(B^T\), which requires \(C\) multiplications and \(C-1\) additions. Since \(A\) has \(N\) rows and \(B^T\) has \(M\) columns, the total computational complexity is:
\begin{equation}
    \begin{aligned}
        F_\text{mat\_mul} & =N\times M\times (C+(C-1)) \\
        & =  N\times M\times (2C-1)
    \end{aligned}
    \label{eq:flops:matrix_multiplication}
\end{equation}

\bd{The FLOPs of a Multi-head Attention}.
Given the input \( Q \in \mathbb{R}^{N_q \times E} \), \( K \in \mathbb{R}^{N_k \times E} \), and \( V \in \mathbb{R}^{N_k \times E} \), a multi-head attention module with \(H\) heads includes the operations shown in \cref{eq:mha:op:qkv_proj}-\cref{eq:mha:op:attn_out}.

In \cref{eq:mha:op:qkv_proj}, the operations involve matrix multiplications: one between \( Q \in \mathbb{R}^{N_q \times E} \) and \( \mathbb{R}^{E \times E} \), and two between \( K, V \in \mathbb{R}^{N_k \times E} \) and \( \mathbb{R}^{E \times E} \), with a total FLOPs:
\begin{equation}
    \begin{aligned}
        N_k\cdot(4E^2-2E)+2N_qE^2-N_qE
    \end{aligned}
    \label{eq:flops:eq_mha_op_qkv_proj}
\end{equation}

In \cref{eq:mha:op:attn_weights}, where $q^h\in\mathbb R^{N_q\times E_h}, k^h\in\mathbb R^{N_k\times E_h}$:

1) It first calculates a matrix multiplication \(q^h\times (k^h)^T\) with FLOPs \(N_qN_k(2E_h-1)\). For total $H$ heads, FLOPs are:
\begin{equation}
    \begin{aligned}
        N_k\cdot (2N_qE-N_qH)
    \end{aligned}
\end{equation}

2) Next, a square root is performed to get \(\sqrt{E}\) with FLOPs $1$.

3) Each element of \(\left(q^h\times (k^h)^T\right)\) divides \(\sqrt{E}\), resulting in FLOPs of $N_qN_k$. For total $H$ heads, FLOPs are:
\begin{equation}
    \begin{aligned}
        N_k\cdot N_qH
    \end{aligned}
\end{equation}

4) For the \(\text{Softmax}\) operation with a vector of shape \(\mathbb R^{N}\) as input, it performs $N$ exponentiations, $N-1$ additions and $N$ divisions. Hence, $\text{Softmax}\left(\cdot\right)$ in \cref{eq:mha:op:attn_weights} has $N_q(3N_k-1)$ FLOPs. For total $H$ heads, FLOPs are:
\begin{equation}
    \begin{aligned}
        N_k\cdot 3N_q H-N_qH
    \end{aligned}
\end{equation}

5) Next, \(O^h=A^h\times v^h\) has \(N_qE_h(2N_k-1)\) FLOPs. For total $H$ heads, FLOPs are:
\begin{equation}
    \begin{aligned}
        N_k\cdot 2N_qE-N_qE
    \end{aligned}
\end{equation}

Hence, \cref{eq:mha:op:attn_weights} and \(O^h=A^h\times v^h\) has FLOPs:
\begin{equation}
    \begin{aligned}
         N_k\cdot(4N_qE+3N_qH)-N_qH-N_q E + 1 
    \end{aligned}
    \label{eq:flops:eq_mha_op_attn_weights_and_oh}
\end{equation}

Finally, \cref{eq:mha:op:attn_out} has FLOPs:
\begin{equation}
    \begin{aligned}
        2N_q E^2-N_q E
    \end{aligned}
    \label{eq:flops:eq_mha_op_attn_out}
\end{equation}

In summary, according to \cref{eq:flops:eq_mha_op_qkv_proj}, \cref{eq:flops:eq_mha_op_attn_weights_and_oh} and \cref{eq:flops:eq_mha_op_attn_out}, a single multi-head attention module with $H$ heads contains FLOPs as follows:
\begin{equation}
    \begin{aligned}
        F_\text{mha}(N_k) & = \lambda N_k + b \\
             \lambda & =4E^2-2E+4N_qE+3N_qH \\
                   b & = 4N_qE^2-3N_qE-N_qH+1
    \end{aligned}
    \label{eq:flops:mha}
\end{equation}

A transformer layer contains a self-attention module, a cross-attention module, two layer normalizations and a feed-forward network.
Among these, only cross-module is related to \textit{key} with $Q\in\mathbb R^{N_q\times E}, K, V\in\mathbb R^{N_k\times E}$ as inputs.
Hence, its FLOPs are:
\begin{equation}
    \begin{aligned}
        F_\text{CA}(N_k)=F_\text{mha}(N_k)
    \end{aligned}
    \label{eq:flops:ca}
\end{equation}

\bd{Cost of Computing Importance Scores}.
To calculate importance scores, we first need to calculate averaged attention maps $A$, which involves $N_qN_k(H-1)$ additions and $N_qN_k$ divisions, with total FLOPs \(N_qN_kH\).
Next, we calculate \(S_0=A\odot\tilde C\), which contains $N_q\times N_k$ multiplications.
After selection, we compute the sum along the column of $S_1$(\cref{eq:sum}), which needs $N_k(k-1)$ additions.
In total, the FLOPs of calculating importance scores are:
\begin{equation}
    \begin{aligned}
        F_\text{S} = N_qN_k H+N_qN_k+N_k(k-1)
    \end{aligned}
    \label{eq:flops:calc_importance_scores}
\end{equation}

Considering a transformer module with $L$ layers applying \method{} with $r, n, k$.

The FLOPs before pruning:
\begin{equation}
    \begin{aligned}
        \text{F}_\text{before} = L\cdot F_\text{CA}(N_k)
    \end{aligned}
    \label{eq:flops:transformer_layer_before_pruning}
\end{equation}

The FLOPs after pruning:
\begin{equation}
    \begin{aligned}
        \text{F}_\text{after} = & \sum\limits_{i=0}^n F_\text{CA}\left(N_k-i\cdot\cfrac r n\right) \\
        & + \max(0, L-(n+1))\cdot F_\text{CA}(N_k-r) \\
        & + \sum\limits_{i=0}^{n-1} F_\text{S}\left( N_k-i\cdot \cfrac{r}{n} \right) 
    \end{aligned}
    \label{eq:flops:transformer_layer_after_pruning}
\end{equation}

\begin{table}[t]
\resizebox{\linewidth}{!}{
\begin{tabular}{c@{\hspace{5pt}}c@{\hspace{5pt}}c|c@{\hspace{5pt}}c|c@{\hspace{5pt}}c}
\toprule
Model                       & Backbone                   & ImageSize                 & $N_k$                    & r     & GFLOPs & Reduced FLOPs\\ \midrule
\multirow{2}{*}{StreamPETR} & \multirow{2}{*}{VovNet}    & \multirow{2}{*}{1600x640} & \multirow{2}{*}{24000} & -     & 174.91 & -             \\
                            &                            &                           &                        & 21000 & 61.44  & -64.88\%      \\ \midrule
\multirow{2}{*}{OPEN}       & \multirow{2}{*}{ResNet101} & \multirow{2}{*}{1408x512} & \multirow{2}{*}{16896} & -     & 123.55 & -             \\
                            &                            &                           &                        & 12000 & 58.84  & -52.37\%      \\
\bottomrule
\end{tabular}
}
\caption{FLOPs reduced of the transformer module after pruning. We show the results of StreamPETR and OPEN with $n=2$, $k=175$ for both models. $1$ GFLOPs = $10^9$ FLOPs.}
\label{tab:flops_before_vs_after_pruning}
\end{table}

\begin{figure}
    \centering
    \includegraphics[width=\linewidth]{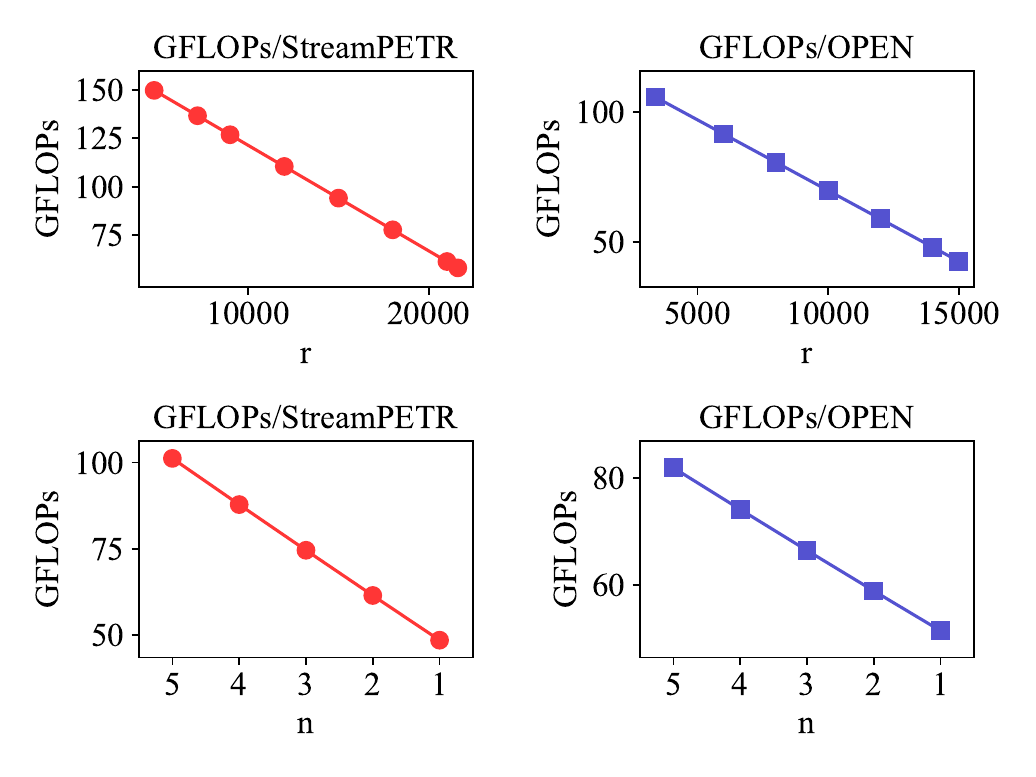}
    \caption{FLOPs decrease with the increase $r$ and the decrease of $n$.
    StreamPETR (OPEN) shown in the figure uses VovNet (ResNet101) and an image size of $1600\times 640$ ($1408\times 512$).
    For the first line, both models use $n$=2 and $k=175$.
    For the second line, StreamPETR (OPEN) uses $k=175$ and $r$=21000 (12000).
    }
    \label{fig:flops_along_r_and_n}
\end{figure}

For different models, we calculate FLOPs before and after pruning as exhibited in \cref{tab:flops_before_vs_after_pruning}.

According to \cref{eq:flops:transformer_layer_after_pruning}, the FLOPs decrease linearly with the increase of $r$ and the decrease of $n$, as shown in \cref{fig:flops_along_r_and_n}.

\begin{figure}
    \centering
    \includegraphics[width=\linewidth]{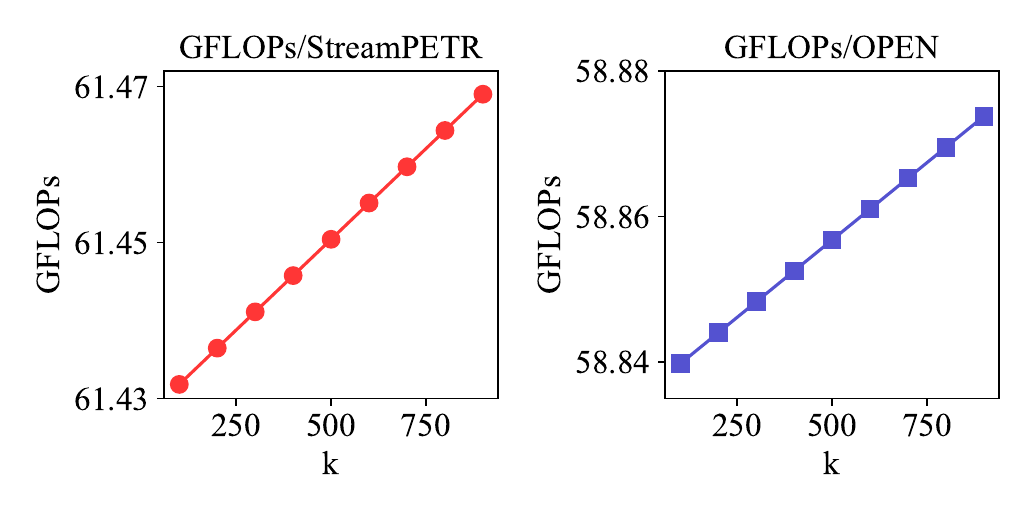}
    \caption{
    When \( k \) changes, there is almost no change in FLOPs.
    }
    \label{fig:flops_along_k}
\end{figure}

As described in \cref{sec:experiments:ablation}, the impact of $k$ on inference time is negligible.
This is because when $k$ varies within the range $[1, N_q]$, its effect on FLOPs is minimal, as shown in \cref{fig:flops_along_k}.
No matter how $k$ changes, the computational cost of the Cross-Attention module varies by no more than $0.1$ GFLOPs.
Compared to the effects of $r$ and $n$, the influence of $k$ on inference time can be ignored.

\begin{table*}[ht]
\centering
\resizebox{0.9\textwidth}{!}{
\begin{tabular}{
    cc@{\hspace{5pt}}c|
    c@{\hspace{5pt}}c@{\hspace{5pt}}c@{\hspace{5pt}}c|
    c@{\hspace{5pt}}c|
    c@{\hspace{5pt}}c@{\hspace{5pt}}c@{\hspace{5pt}}c@{\hspace{5pt}}c|c
}
\toprule
\bd{Model}
& \bd{Backbone}
& \bd{ImageSize}
& $N_k$
& \bd{$r$}
& \bd{$n$}
& \bd{$k$}
& \bd{mAP} $\uparrow$
& \bd{NDS} $\uparrow$
& \bd{mATE} $\downarrow$
& \bd{mASE} $\downarrow$
& \bd{mAOE} $\downarrow$
& \bd{mAVE} $\downarrow$
& \bd{mAAE} $\downarrow$
& \bd{Inf. Time(ms)} $\downarrow$ \\ \midrule
\ml{2}{PETR}       & \ml{2}{ResNet50} & \ml{2}{1408x512}& \ml{2}{16896}& -     & - & -   & 31.74\% & 0.3669 & 0.8392 & 0.2797 & 0.6145 & 0.9521 & 0.2322 & 131.94\\
                   &                  &                 &              & 10000 & 1 & 175 & 30.78\% & 0.3579 & 0.8540 & 0.2832 & 0.6168 & 0.9703 & 0.2354 & 110.35(-16.36\%)\\ \midrule
\ml{2}{PETRv2}     & \ml{2}{VovNet}   & \ml{2}{800x320} & \ml{2}{12000}& -     & - & -   & 41.05\% & 0.5024 & 0.7232 & 0.2692 & 0.4529 & 0.3896 & 0.1932 & 157.52\\
                   &                  &                 &              & 8000  & 2 & 175 & 40.29\% & 0.4918 & 0.7333 & 0.2750 & 0.4526 & 0.4428 & 0.1930 & 139.33(-11.55\%)\\ \midrule
\ml{2}{StreamPETR} & \ml{2}{ResNet50} & \ml{2}{704x256} & \ml{2}{4224} & -     & - & -   & 38.01\% & 0.4822 & 0.6781 & 0.2763 & 0.6401 & 0.2831 & 0.2007 & 60.52 \\
                   &                  &                 &              & 2000  & 1 & 900 & 37.96\% & 0.4822 & 0.6828 & 0.2757 & 0.6373 & 0.2794 & 0.2009 & 51.59(-14.76\%)\\ \midrule
\ml{2}{3DPPE}      & \ml{2}{VovNet}   & \ml{2}{800x320} & \ml{2}{6000} & -     & - & -   & 39.81\% & 0.4460 & 0.7040 & 0.2699 & 0.4951 & 0.8438 & 0.2177 & 125.13 \\
                   &                  &                 &              & 3000  & 1 & 175 & 39.57\% & 0.4430 & 0.7089 & 0.2729 & 0.4974 & 0.8496 & 0.2201 & 94.09(-24.81\%)\\ \midrule
\ml{2}{M-BEV}      & \ml{2}{VovNet}   & \ml{2}{800x320} & \ml{2}{12000}& -     & - & -   & 35.14\% & 0.4640 & 0.7300 & 0.2717 & 0.4980 & 0.4324 & 0.1845 & 178.87 \\
                   &                  &                 &              & 6000  & 1 & 175 & 34.91\% & 0.4600 & 0.7379 & 0.2727 & 0.5049 & 0.4428 & 0.1873 & 150.93(-15.63\%)\\ \midrule
\ml{6}{OPEN}       & \ml{2}{ResNet50} & \ml{2}{704x256} & \ml{2}{4224} & -     & - & -   & 47.02\% & 0.5657 & 0.5676 & 0.2702 & 0.4221 & 0.2321 & 0.2019 & 77.25 \\
                   &                  &                 &              & 2000  & 1 & 900 & 46.88\% & 0.5636 & 0.5687 & 0.2721 & 0.2332 & 0.2315 & 0.2026 & 63.31(-18.05\%)\\ \cline{2-15}
                   & \ml{2}{VovNet}   & \ml{2}{800x320} & \ml{2}{6000} & -     & - & -   & 52.07\% & 0.6128 & 0.5250 & 0.2566 & 0.2811 & 0.2148 & 0.1982 & 118.55 \\
                   &                  &                 &              & 3000  & 1 & 175 & 52.12\% & 0.6130 & 0.5250 & 0.2569 & 0.2810 & 0.2148 & 0.1985 & 110.21(-7.04\%)\\ \cline{2-15}
                   & \ml{2}{ResNet101}& \ml{2}{1408x512}& \ml{2}{16896}& -     & - & -   & 51.80\% & 0.6043 & 0.5314 & 0.2679 & 0.3457 & 0.2095 & 0.1922 & 196.55 \\
                   &                  &                 &              & 12000 & 1 & 175 & 51.47\% & 0.6018 & 0.5356 & 0.2691 & 0.3446 & 0.2131 & 0.1931 & 174.70(-11.16\%)\\ \midrule
\ml{2}{ToC3D}      & \ml{2}{ToC3DViT} & \ml{2}{1600x800}& \ml{2}{30000}& -     & - & -   & 54.20\% & 0.6187 & 0.5589 & 0.2571 & 0.2716 & 0.2353 & 0.2007 & 863.47 \\
                   &                  &                 &              & 27000 & 1 & 900 & 53.31\% & 0.6121 & 0.5736 & 0.2591 & 0.2713 & 0.2381 & 0.2029 & 809.48(-6.25\%)\\
\bottomrule
\end{tabular}
}
\caption{Optimal results for each model.}
\label{tab:optimal_results}
\end{table*}

\begin{table*}[ht]
\centering
\resizebox{0.9\textwidth}{!}{
\begin{tabular}{cc@{\hspace{5pt}}c|c@{\hspace{5pt}}c@{\hspace{5pt}}c|c@{\hspace{5pt}}c|c@{\hspace{5pt}}c@{\hspace{5pt}}c@{\hspace{5pt}}c@{\hspace{5pt}}c}
\toprule
  \textbf{Model} 
& \textbf{Backbone}
& \textbf{ImageSize}
& \textbf{Pruning}
& $r$
& $n$
& \textbf{mAP} $\uparrow$
& \textbf{NDS} $\uparrow$
& \textbf{mATE} $\downarrow$
& \textbf{mASE} $\downarrow$
& \textbf{mAOE} $\downarrow$
& \textbf{mAVE} $\downarrow$
& \textbf{mAAE} $\downarrow$ \\ \midrule
\multirow{14}{*}{PETR}
      & \multirow{7}{*}{ResNet50}    & \multirow{7}{*}{1408x512}
      & -   & -     & - & 31.74\% & 0.3669 & 0.8392 & 0.2797 & 0.6145 & 0.9521 & 0.2322 \\ \cmidrule{4-13}
& & & \ml{3}{ToMe}  & 8000  & 2 & 30.48\% & 0.3543 & 0.8708 & 0.2816 & 0.6091 & 0.9852 & 0.2342 \\
& & &               & 8000  & 1 & 29.81\% & 0.3490 & 0.8714 & 0.2823 & 0.6173 & 0.9878 & 0.2418 \\
& & &               & 12000 & 2 & 29.63\% & 0.3461 & 0.8859 & 0.2836 & 0.6198 & 0.9895 & 0.2412 \\  \cmidrule{4-13}
& & & \ml{3}{Ours}  & 8000  & 2 & 31.22\% & 0.3639 & 0.8436 & 0.2808 & 0.6133 & 0.9517 & 0.2329 \\
& & &               & 8000  & 1 & 31.58\% & 0.3651 & 0.8418 & 0.2801 & 0.6148 & 0.9576 & 0.2330 \\
& & &               & 12000 & 2 & 30.78\% & 0.3579 & 0.8540 & 0.2832 & 0.6168 & 0.9703 & 0.2354 \\ \cmidrule{2-13}
      & \multirow{7}{*}{VovNet}    & \multirow{7}{*}{1600x640}
      & -   & -     & - & 40.45\% & 0.4517 & 0.7287 & 0.2706 & 0.4485 & 0.8399 & 0.2178 \\ \cmidrule{4-13}
& & & \ml{3}{ToMe}  & 12000 & 2 & 21.88\% & 0.2835 & 0.9517 & 0.4471 & 0.5913 & 1.0230 & 0.2688 \\
& & &               & 12000 & 1 & 27.17\% & 0.3350 & 0.8710 & 0.3560 & 0.5512 & 0.9959 & 0.2342 \\
& & &               & 18000 & 2 & 26.58\% & 0.3162 & 0.9143 & 0.4300 & 0.5663 & 1.0210 & 0.2564 \\ \cmidrule{4-13}
& & & \ml{3}{Ours}  & 12000 & 2 & 40.42\% & 0.4502 & 0.7305 & 0.2702 & 0.4501 & 0.8512 & 0.2172 \\
& & &               & 12000 & 1 & 39.37\% & 0.4425 & 0.7429 & 0.2722 & 0.4592 & 0.8517 & 0.2174 \\
& & &               & 18000 & 2 & 39.53\% & 0.4432 & 0.7482 & 0.2720 & 0.4538 & 0.8539 & 0.2167 \\ \midrule
\multirow{5}{*}{3DPPE}
      & \multirow{5}{*}{VovNet}    & \multirow{5}{*}{800x320}
    & -             & -            & - & 39.81\% & 0.4460 & 0.7040 & 0.2699 & 0.4951 & 0.8438 & 0.2177 \\ \cmidrule{4-13}
& & & \ml{2}{ToMe}  & \ml{2}{2000} & 2 & 39.56\% & 0.4432 & 0.7083 & 0.2715 & 0.4972 & 0.8490 & 0.2198 \\
& & &               &              & 1 & 36.99\% & 0.4197 & 0.7500 & 0.2719 & 0.5337 & 0.8735 & 0.2233 \\  \cmidrule{4-13}
& & & \ml{2}{Ours}  & \ml{2}{2000} & 2 & 39.74\% & 0.4449 & 0.7057 & 0.2707 & 0.4956 & 0.8465 & 0.2202 \\ 
& & &               &              & 1 & 39.57\% & 0.4430 & 0.7089 & 0.2729 & 0.4974 & 0.8496 & 0.2201 \\ \midrule
\multirow{5}{*}{MV2D}
    & \multirow{5}{*}{ResNet50} & \multirow{5}{*}{1408x512} 
    & -             & -            & - & 44.92\% & 0.5399 & 0.6246 & 0.2657 & 0.3840 & 0.4009 & 0.1722 \\ \cmidrule{4-13}
& & & \ml{2}{ToMe}  & \ml{2}{50\%} & 2 & 13.62\% & 0.2754 & 0.9028 & 0.3472 & 0.7737 & 0.6777 & 0.2261 \\
& & &               &              & 1 & 13.45\% & 0.2721 & 0.9000 & 0.3477 & 0.7890 & 0.6864 & 0.2286 \\ \cmidrule{4-13}
& & & \ml{2}{Ours}  & \ml{2}{50\%} & 2 & 44.11\% & 0.5384 & 0.6248 & 0.2657 & 0.3844 & 0.4024 & 0.1717 \\
& & &               &              & 1 & 41.17\% & 0.5183 & 0.6289 & 0.2693 & 0.3887 & 0.4165 & 0.1717 \\
\bottomrule
\end{tabular}
}
\caption{
More results of comparison to ToMe~\cite{tome}.
Due to the use of bipartite matching, ToMe cannot prune more than 50\% of keys in one layer; hence, some results of $r>N_k/2$ are lacking.
}
\label{tab:results_more_comparison}
\end{table*}

\section{More Results}
\label{sec:appendix:more_reuslts}

\subsection{Optimal Results}
\label{sec:appendix:more_results:optimal_results}

As described in \cref{sec:experiments:results_performance}, to facilitate clearer comparison and save space, we uniformly report the results with \( n = 2 \) and \( k = 175 \).
However, this is not the optimal configuration for every model.
Here, the optimal configuration refers to the maximum value of \( r \) and the minimum value of \( n \) that can be applied when the mAP and NDS decrease by no more than 1\%.
For the vast majority of models, we can directly use \( n = 1 \) for pruning, as shown in \cref{tab:optimal_results}.
For models that have already achieved optimal results, as shown in \cref{tab:results-performance} (e.g., PETR-vov, FocalPETR, MV2D, StreamPETR-vov), we do not repeat the results here.

\begin{table*}[ht]
\resizebox{\textwidth}{!}{
\begin{tabular}{ccc|ccc|cc|ccccc}
\toprule
  \textbf{Model}
& \textbf{Backbone}
& \textbf{ImageSize}
& $r$
& $n$
& $k$
& \textbf{mAP} $\uparrow$
& \textbf{NDS} $\uparrow$
& \textbf{mATE} $\downarrow$
& \textbf{mASE} $\downarrow$
& \textbf{mAOE} $\downarrow$
& \textbf{mAVE} $\downarrow$
& \textbf{mAAE} $\downarrow$ \\ \midrule
\multirow{22}{*}{StreamPETR} & \multirow{6}{*}{ResNet50} & \multirow{6}{*}{704x256}
     & -            & - & -   & 38.01\% & 0.4822 & 0.6781 & 0.2763 & 0.6401 & 0.2831 & 0.2007 \\ \cmidrule{4-13}
 & & & \ml{5}{2000} & 5 & 150 & 38.04\% & 0.4824 & 0.6787 & 0.2760 & 0.6388 & 0.2838 & 0.2007 \\
 & & &              & 4 & 175 & 38.04\% & 0.4826 & 0.6795 & 0.2761 & 0.6367 & 0.2826 & 0.2014 \\
 & & &              & 3 & 150 & 37.91\% & 0.4813 & 0.6819 & 0.2761 & 0.6385 & 0.2832 & 0.2025 \\
 & & &              & 2 & 175 & 37.91\% & 0.4817 & 0.6787 & 0.2758 & 0.6390 & 0.2844 & 0.2016 \\
 & & &              & 1 & 900 & 37.96\% & 0.4822 & 0.6828 & 0.2757 & 0.6373 & 0.2794 & 0.2009 \\ \cmidrule{2-13}
 & \multirow{16}{*}{VovNet} & \multirow{16}{*}{1600x640}
     & -           & - & - & 48.89\% & 0.5732 & 0.6096 & 0.2601 & 0.3882 & 0.2603 & 0.1944 \\ \cmidrule{4-13}
 &&& \ml{5}{12000} & 5 & 900 & 48.92\% & 0.5734 & 0.6089 & 0.2603 & 0.3884 & 0.2601 & 0.1942 \\
 &&&               & 4 & 900 & 48.90\% & 0.5735 & 0.6082 & 0.2601 & 0.3871 & 0.2598 & 0.1946 \\
 &&&               & 3 & 900 & 48.89\% & 0.5736 & 0.6076 & 0.2599 & 0.3870 & 0.2595 & 0.1940 \\
 &&&               & 2 & 900 & 48.95\% & 0.5741 & 0.6074 & 0.2605 & 0.3858 & 0.2594 & 0.1937 \\
 &&&               & 1 & 175 & 48.85\% & 0.5738 & 0.6078 & 0.2603 & 0.3813 & 0.2613 & 0.1941 \\ \cmidrule{4-13}
 &&& \ml{5}{18000} & 5 & 900 & 48.95\% & 0.5743 & 0.6059 & 0.2605 & 0.3833 & 0.2602 & 0.1944 \\
 &&&               & 4 & 900 & 48.94\% & 0.5744 & 0.6064 & 0.2604 & 0.3821 & 0.2600 & 0.1943 \\
 &&&               & 3 & 900 & 48.67\% & 0.5738 & 0.6054 & 0.2617 & 0.3701 & 0.2627 & 0.1955 \\
 &&&               & 2 & 900 & 48.85\% & 0.5738 & 0.6054 & 0.2617 & 0.3701 & 0.2627 & 0.1955 \\
 &&&               & 1 & 900 & 48.62\% & 0.5734 & 0.6087 & 0.2613 & 0.3705 & 0.2628 & 0.1942 \\ \cmidrule{4-13}
 &&& \ml{5}{21000} & 5 & 900 & 48.92\% & 0.5742 & 0.6063 & 0.2604 & 0.3820 & 0.2611 & 0.1942 \\
 &&&               & 4 & 900 & 48.85\% & 0.5756 & 0.6015 & 0.2614 & 0.3659 & 0.2616 & 0.1966 \\
 &&&               & 3 & 175 & 48.21\% & 0.5702 & 0.6089 & 0.2619 & 0.3751 & 0.2669 & 0.1957 \\
 &&&               & 2 & 900 & 48.71\% & 0.5731 & 0.6052 & 0.2622 & 0.3822 & 0.2618 & 0.1933 \\
 &&&               & 1 & 900 & 47.87\% & 0.5695 & 0.6071 & 0.2635 & 0.3625 & 0.2692 & 0.1952 \\ \midrule
\multirow{22}{*}{OPEN} & \multirow{6}{*}{ResNet50} & \multirow{6}{*}{704x256}
     & -            & - & -   & 47.02\% & 0.5657 & 0.5676 & 0.2702 & 0.4221 & 0.2321 & 0.2019 \\ \cmidrule{4-13}
 & & & \ml{5}{2000} & 5 & 900 & 46.98\% & 0.5648 & 0.5669 & 0.2706 & 0.4303 & 0.2319 & 0.2015 \\
 & & &              & 4 & 900 & 47.03\% & 0.5649 & 0.5680 & 0.2705 & 0.4296 & 0.2321 & 0.2017 \\
 & & &              & 3 & 900 & 47.02\% & 0.5642 & 0.5685 & 0.2706 & 0.4345 & 0.2324 & 0.2028 \\
 & & &              & 2 & 175 & 46.85\% & 0.5637 & 0.5682 & 0.2705 & 0.4311 & 0.2325 & 0.2031 \\
 & & &              & 1 & 900 & 46.88\% & 0.5636 & 0.5687 & 0.2721 & 0.4332 & 0.2315 & 0.2026 \\ \cmidrule{2-13}
 & \multirow{6}{*}{VovNet} & \multirow{6}{*}{800x320}
     & -           & - &  -  & 52.07\% & 0.6128 & 0.5250 & 0.2566 & 0.2811 & 0.2148 & 0.1982 \\ \cmidrule{4-13}
 &&& \ml{5}{3000}  & 5 & 175 & 52.07\% & 0.6126 & 0.5261 & 0.2566 & 0.2818 & 0.2147 & 0.1985 \\
 &&&               & 4 & 175 & 52.06\% & 0.6124 & 0.5267 & 0.2566 & 0.2823 & 0.2144 & 0.1987 \\
 &&&               & 3 & 175 & 52.08\% & 0.6127 & 0.5254 & 0.2565 & 0.2814 & 0.2149 & 0.1986 \\
 &&&               & 2 & 175 & 52.09\% & 0.6129 & 0.5249 & 0.2569 & 0.2808 & 0.2146 & 0.1986 \\
 &&&               & 1 & 175 & 52.12\% & 0.6130 & 0.5250 & 0.2569 & 0.2810 & 0.2148 & 0.1985 \\ \cmidrule{2-13}
 & \multirow{11}{*}{ResNet101} & \multirow{11}{*}{1408x512}
     & -           & - &  -  & 51.80\% & 0.6043 & 0.5314 & 0.2679 & 0.3457 & 0.2095 & 0.1922 \\ \cmidrule{4-13}
 &&& \ml{5}{10000} & 5 & 900 & 51.78\% & 0.6033 & 0.5339 & 0.2694 & 0.3509 & 0.2101 & 0.1916 \\
 &&&               & 4 & 900 & 51.75\% & 0.6029 & 0.5333 & 0.2693 & 0.3544 & 0.2109 & 0.1907 \\
 &&&               & 3 & 900 & 51.58\% & 0.6021 & 0.5372 & 0.2695 & 0.3496 & 0.2107 & 0.1914 \\
 &&&               & 2 & 900 & 51.63\% & 0.6017 & 0.5386 & 0.2713 & 0.3542 & 0.2102 & 0.1906 \\
 &&&               & 1 & 900 & 51.68\% & 0.6032 & 0.5357 & 0.2691 & 0.3443 & 0.2110 & 0.1921 \\ \cmidrule{4-13}
 &&& \ml{5}{12000} & 5 & 900 & 51.75\% & 0.6030 & 0.5346 & 0.2698 & 0.3525 & 0.2109 & 0.1901 \\
 &&&               & 4 & 900 & 51.78\% & 0.6024 & 0.5343 & 0.2700 & 0.3593 & 0.2114 & 0.1900 \\
 &&&               & 3 & 900 & 51.40\% & 0.5991 & 0.5425 & 0.2712 & 0.3634 & 0.2113 & 0.1902 \\
 &&&               & 2 & 175 & 51.57\% & 0.6019 & 0.5368 & 0.2726 & 0.3493 & 0.2102 & 0.1905 \\
 &&&               & 1 & 175 & 51.47\% & 0.6018 & 0.5356 & 0.2691 & 0.3446 & 0.2131 & 0.1931 \\
\bottomrule
\end{tabular}
}
\caption{More results with different $r$, $n$ and $k$ for StreamPETR and OPEN.}
\label{tab:results_more_ablation_rnk}
\end{table*}

\subsection{More Comparison to ToMe}
\label{sec:appendix:more_results:more_comparison}

As described in \cref{sec:experiments:comparison}, our method is the first to apply zero-shot pruning to 3D object detection models. Therefore, there are no similar methods available for a fair comparison. However, we can attempt to transfer zero-shot pruning methods from the Vision Transformer (ViT) domain to 3D object detection models. Among these methods, ATS relies on a classification token specifically designed for classification tasks, and Zero-TPrune depends on a square-shaped attention map. Neither of these features exists in 3D object detection methods, making it impossible to transfer ATS and Zero-TPrune to 3D object detection. In contrast, ToMe does not have a strong dependency on the shape of the attention map or the classification token. Therefore, we selected ToMe as the comparison method.

As shown in the \cref{tab:results_more_comparison}, for PETR and MV2D, ToMe performs extremely poorly---when pruning only 50\% of the keys, the model's performance completely collapses, with mAP dropping by more than 10\%.
In contrast, \method{} can maintain model performance.
Even when $n=1$ and some results are suboptimal, with mAP dropping by more than 1\%, the model does not degrade into an unusable state.

\subsection{More Ablation Experiments}
\label{sec:appendix:more_results:more_ablation_exps}



We report additional results for different values of \( r, n, k \) in \cref{tab:results_more_ablation_rnk}.
As discussed in the main text \cref{sec:experiments:results_performance}, the optimal parameter selection varies across different models.  
If we set a 1\% mAP drop as the threshold, the maximum number of keys that can be pruned varies depending on the model, and the same applies to \( n \). When \( r \) is too large or \( n = 1 \), some models may experience a performance drop exceeding 1\%.
However, the model performance does not completely degrade into an unusable state but remains at a functional level.

We have conducted extensive experiments for each model, but due to space limitations, only a portion of the results can be presented here.
Please refer to our GitHub repository for additional experimental results.

\section{Further Exploration}
\label{sec:appendix:further_exploration}

\begin{table*}[ht]
\centering
\resizebox{0.9\textwidth}{!}{%
\begin{tabular}{c|cc|cc|ccccc|c}
\toprule
  \bd{Model}
& \bd{r}
& \bd{q}
& \bd{mAP} $\uparrow$
& \bd{NDS} $\uparrow$
& \bd{mATE} $\downarrow$
& \bd{mASE} $\downarrow$
& \bd{mAOE} $\downarrow$
& \bd{mAVE} $\downarrow$
& \bd{mAAE} $\downarrow$
& \bd{Dec. Time (ms)} \\ \midrule
\multirow{4}{*}{StreamPETR} & -                      & -   & 48.89\% & 0.5732 & 0.6096 & 0.2601 & 0.3882 & 0.2603 & 0.1944 & 64.93          \\ \cmidrule{2-11}
                            & \multirow{3}{*}{21000} & -   & 48.55\% & 0.5730 & 0.6033 & 0.2626 & 0.3771 & 0.2611 & 0.1941 & 34.98          \\
                            &                        & 300 & 48.42\% & 0.5703 & 0.6055 & 0.2633 & 0.3912 & 0.2620 & 0.1958 & 34.09          \\
                            &                        & 600 & 47.46\% & 0.5606 & 0.6239 & 0.2661 & 0.3952 & 0.2826 & 0.1992 & 29.42          \\ \midrule
\multirow{4}{*}{OPEN}       & -                      & -   & 51.80\% & 0.6043 & 0.5314 & 0.2679 & 0.3457 & 0.2095 & 0.1922 & 46.39          \\ \cmidrule{2-11}
                            & \multirow{3}{*}{12000} & -   & 51.57\% & 0.6019 & 0.5368 & 0.2726 & 0.3493 & 0.2102 & 0.1905 & 28.99          \\
                            &                        & 300 & 51.48\% & 0.6014 & 0.5389 & 0.2727 & 0.3460 & 0.2088 & 0.1939 & 26.31          \\
                            &                        & 600 & 50.24\% & 0.5922 & 0.5542 & 0.2732 & 0.3524 & 0.2144 & 0.1959 & 25.97          \\
\bottomrule
\end{tabular}%
}
\caption{Results of pruning queries. We use StreamPETR-vov-1600x640 and OPEN-r101-1408x512.}
\label{tab:results_pruning_queries}
\end{table*}

\begin{table}[t]
\centering
\resizebox{\linewidth}{!}{
\begin{tabular}{c|cc|ccccc}
\toprule
\textbf{r} 
& \textbf{mAP} $\uparrow$
& \textbf{NDS} $\uparrow$
& \textbf{mATE} $\downarrow$
& \textbf{mASE} $\downarrow$
& \textbf{mAOE} $\downarrow$
& \textbf{mAVE} $\downarrow$
& \textbf{mAAE} $\downarrow$ \\ \midrule
-      & 43.07\% & 0.5389 & 0.6023 & 0.2686 & 0.4238 & 0.2597 & 0.2105 \\
2000   & 42.99\% & 0.5382 & 0.6035 & 0.2696 & 0.4230 & 0.2609 & 0.2099 \\
\bottomrule
\end{tabular}
}
\caption{
Pruning fully converged models.
We train StreamPETR-r50-704x256 for 120 epochs to ensure its full convergence.
The first line remaining $r$ blank is the original model's results without pruning.
}\label{tab:results_fully_converged}
\end{table}

\subsection{Why Some Models Improve in Performance}
\label{sec:appendix:analysis:reason_for_performance_improvement}

We can observe that for some models, the mAP and NDS increase rather than decrease after pruning, such as FocalPETR-vov-800x320 in \cref{tab:results-performance} and StreamPETR-r50-704x256 with $n$=4 in \cref{tab:results_more_ablation_rnk}.
We believe this is related to the redundant information in the image features. As shown in \cref{fig:3d_detector}, the key is the image feature extracted by the backbone, which inevitably contains background information (such as sky, buildings, etc.) that is ineffective for object detection.
These keys interact with the query and affect the detection performance.

In the original 3D detectors, for each transformer decoder layer, the query is continuously updated, while the key and value remain unchanged.
Therefore, the background key repeatedly influences the query.
In fact, it can be argued that the ``unimportant keys'' pruned by our method are essentially background tokens.
It is precisely because these keys, which interfere with detection, are pruned that the phenomenon of increased model performance occurs.

\subsection{Pruning Queries}
\label{subsec:appendix:more_results:pruning_queries}

While it is possible to prune both keys and queries at runtime, the latter's involvement in self-attention operations limits the extent of pruning.
To ensure that the mAP degradation does not exceed 1\%, we cannot prune 300 queries, offering only marginal speed improvements.
Conversely, to achieve significant acceleration in model speed, pruning 600 queries would result in a sharp decline in mAP.
The results are shown in \cref{tab:results_pruning_queries}.

We believe that pruning the key is more effective than pruning the query for the following reasons:
The key does not have explicit self-attention. In contrast, after interacting with the key, the query is fed into the self-attention mechanism of the next layer. This introduces internal dependencies, meaning that even if a query generates a low classification score, its value may influence queries with high classification scores through self-attention. Therefore, pruning the query can have a significant impact on the remaining queries, thereby degrading model performance. In contrast, the dependencies of the key are indirect, so pruning the key has a lower impact on model performance.
Moreover, keys contain redundant information more than queries.
Please see the analysis in \cref{sec:appendix:analysis:reason_for_performance_improvement}.

\subsection{Pruning Fully Converged Models}
\label{sec:appendix:more_results:fully_converged_models}

To ensure a fair comparison with prior work while considering training efficiency, many previous experiments use a 24-epoch training schedule, which often does not achieve full convergence. To assess whether \method{} remains effective after full convergence, we trained a StreamPETR-ResNet50-704x256 model for 120 epochs. As shown in \cref{tab:results_fully_converged}, \method{} preserves the model’s performance even after full convergence, with only a 0.08\% decrease in mAP and a 0.0007 reduction in NDS.

\begin{table}[t]
\centering
\resizebox{\linewidth}{!}{
\begin{tabular}{c|c@{\hspace{5pt}}c|c@{\hspace{5pt}}c@{\hspace{5pt}}c@{\hspace{5pt}}c@{\hspace{5pt}}c|c}
\toprule
$r$
& \textbf{mAP} $\uparrow$
& \textbf{NDS} $\uparrow$
& \textbf{mATE} $\downarrow$
& \textbf{mASE} $\downarrow$
& \textbf{mAOE} $\downarrow$
& \textbf{mAVE} $\downarrow$
& \textbf{mAAE} $\downarrow$
& \bd{Training Time}
\\ \midrule
-     & 48.89\% & 0.5732 & 0.6096 & 0.2601 & 0.3882 & 0.2603 & 0.1944 & 2d 14h\\
21000 & 49.42\% & 0.5787 & 0.5982 & 0.2579 & 0.3651 & 0.2698 & 0.1926 & 2d 13h\\
\bottomrule
\end{tabular}
}
\caption{
Training StreamPETR-vov-1600x640 with \method{}, while \method{} is applied, $n$ and $k$ are set to 1 and 175, respectively.
}\label{tab:results_training_with_tggbc}
\end{table}

\begin{table*}[t]
\centering
\resizebox{0.95\textwidth}{!}{
\begin{tabular}{cc|c|ccc|ccc|c}
\toprule
  \bd{Model}
& \bd{Backbone}
& \bd{\Method{}}
& \textbf{$\mathbf{mAP}$} $\uparrow$
& \textbf{$\mathbf{AP_{50}}$} $\uparrow$
& \textbf{$\mathbf{AP_{75}}$} $\uparrow$
& \textbf{$\mathbf{AP_s}$} $\uparrow$
& \textbf{$\mathbf{AP_m}$} $\uparrow$
& \textbf{$\mathbf{AP_l}$} $\uparrow$
& \textbf{Inf. Time (ms)} $\downarrow$\\ \midrule
\ml{4}{DETR}            & \ml{2}{ResNet50}  & -     & 0.421 & 0.623 & 0.442 & 0.214 & 0.460 & 0.610 & 42.06           \\
                        &                   & \ck{} & 0.414 & 0.620 & 0.436 & 0.205 & 0.455 & 0.603 & 35.24(-16.21\%, 1.19x) \\ \cmidrule{2-10}
                        & \ml{2}{ResNet101} & -     & 0.435 & 0.638 & 0.463 & 0.218 & 0.480 & 0.480 & 54.61 \\
                        &                   & \ck{} & 0.426 & 0.635 & 0.453 & 0.211 & 0.470 & 0.608 & 47.01(-13.92\%, 1.16x) \\ \midrule
\ml{4}{ConditionalDETR} & \ml{2}{ResNet50}  & -     & 0.409 & 0.619 & 0.434 & 0.207 & 0.442 & 0.595 & 43.88 \\
                        &                   & \ck{} & 0.400 & 0.615 & 0.427 & 0.196 & 0.436 & 0.587 & 40.13 (-8.55\%, 1.09x)\\ \cmidrule{2-10}
                        & \ml{2}{ResNet101} & -     & 0.428 & 0.636 & 0.459 & 0.218 & 0.467 & 0.610 & 69.97 \\
                        &                   & \ck{} & 0.424 & 0.635 & 0.454 & 0.215 & 0.463 & 0.605 & 55.66 (-20.45\%, 1.26x)\\
\bottomrule
\end{tabular}
}
\caption{Results of DETR and ConditionalDETR with \method{}.
Lines with ``\method{}'' remaining blank are the original results without pruning.
}
\label{tab:results_tggbc_on_2d_object_detection}
\vskip -0.1in
\end{table*}

\subsection{Training with \method{}}
\label{subsec:appendix:more_results:training_with_our_method}

If there is a new model, one can also train it with \method{} from the beginning, as shown in \cref{tab:results_training_with_tggbc}.
Our method is capable of reducing training time.
For example, training StreamPETR-vov-1600x640 with \( r = 21000 \) and \( n = 1 \) for 30 epochs takes less time than training without \method{} for 24 epochs while achieving a better mAP.

\subsection{2D Object Detection Models with \method{}}

As described in \cref{subsec:related_works:pruning_methods}, the number of keys in ViT-based methods is significantly smaller than that in 3D object detection methods.
Similarly, the number of keys in 2D object detection is around 1,000 (e.g., in ConditionalDETR).
This is also why we focus on 3D object detection rather than extensively studying 2D object detection methods.
Moreover, current 2D object detection methods are rapidly evolving and highly mature. Methods based on DETR are not the absolute mainstream, as other approaches, such as the YOLO series, are still widely used in 2D object detection tasks.

Additionally, 3D object detection is a highly practical and valuable task.
Therefore, from the very beginning, we focused on pruning 3D object detection models.
However, for some DETR-based 2D detection methods, \method{} can still be applied.
Here, we take ConditionalDETR as an example to verify the effectiveness of \method{} on 2D object detection methods, as shown in \cref{tab:results_tggbc_on_2d_object_detection}.

In ConditionalDETR, the number of keys is not always the same.
Therefore, we adopt a configuration similar to MV2D, using \( r \) to represent the pruning ratio and set a threshold \( t \). When the current number of keys exceeds \( t \), pruning is performed.
Through experiments, \method{} can reduce the model's inference time by 19.17\% ($1.24\times $) while keeping the mAP degradation below 1\%.

\end{document}